\documentclass[letterpaper, 10 pt, conference]{ieeeconf}  

\IEEEoverridecommandlockouts                              

\usepackage{graphics} 
\usepackage{epsfig} 
\usepackage{times} 
\usepackage{amsmath} 
\usepackage{amssymb}  
\usepackage{microtype}
\usepackage{cite}
\title{\LARGE \bf
Video Depth Estimation by Fusing Flow-to-Depth Proposals
}

\author{Jiaxin Xie, Chenyang Lei, Zhuwen Li, Li Erran Li, and Qifeng Chen
\thanks{Jiaxin Xie (jxieax@connect.ust.hk) and Chenyang Lei (cleiaa@ust.hk) are with the Department of Computer Science and Engineering, HKUST. Zhuwen Li (lzhuwen@gmail.com) is with Nuro Inc. Li Erran Li (erranlli@gmail.com) is with Scale AI. Qifeng Chen (cqf@ust.hk) is with the Department of Computer Science and Engineering and the Department of Electronic and Computer Enginnering, HKUST.  }
}

\begin{document}

\maketitle
\thispagestyle{empty}
\pagestyle{empty}

\begin{abstract}
Depth from a monocular video can enable billions of devices and robots with a single camera to see the world in 3D. In this paper, we present an approach with a differentiable flow-to-depth layer for video depth estimation. The model consists of a flow-to-depth layer, a camera pose refinement module, and a depth fusion network. Given optical flow and camera pose, our flow-to-depth layer generates depth proposals and the corresponding confidence maps by explicitly solving an epipolar geometry optimization problem. Our flow-to-depth layer is differentiable, and thus we can refine camera poses by maximizing the aggregated confidence in the camera pose refinement module. Our depth fusion network can utilize depth proposals and their confidence maps inferred from different adjacent frames to produce the final depth map. Furthermore, the depth fusion network can additionally take the depth proposals generated by other methods to improve the results further. The experiments on three public datasets show that our approach outperforms state-of-the-art depth estimation methods, and has reasonable cross dataset generalization capability: our model trained on KITTI still performs well on the unseen Waymo dataset.

\end{abstract}

\begin{figure*}[h!]
\centering
\includegraphics[width=\linewidth]{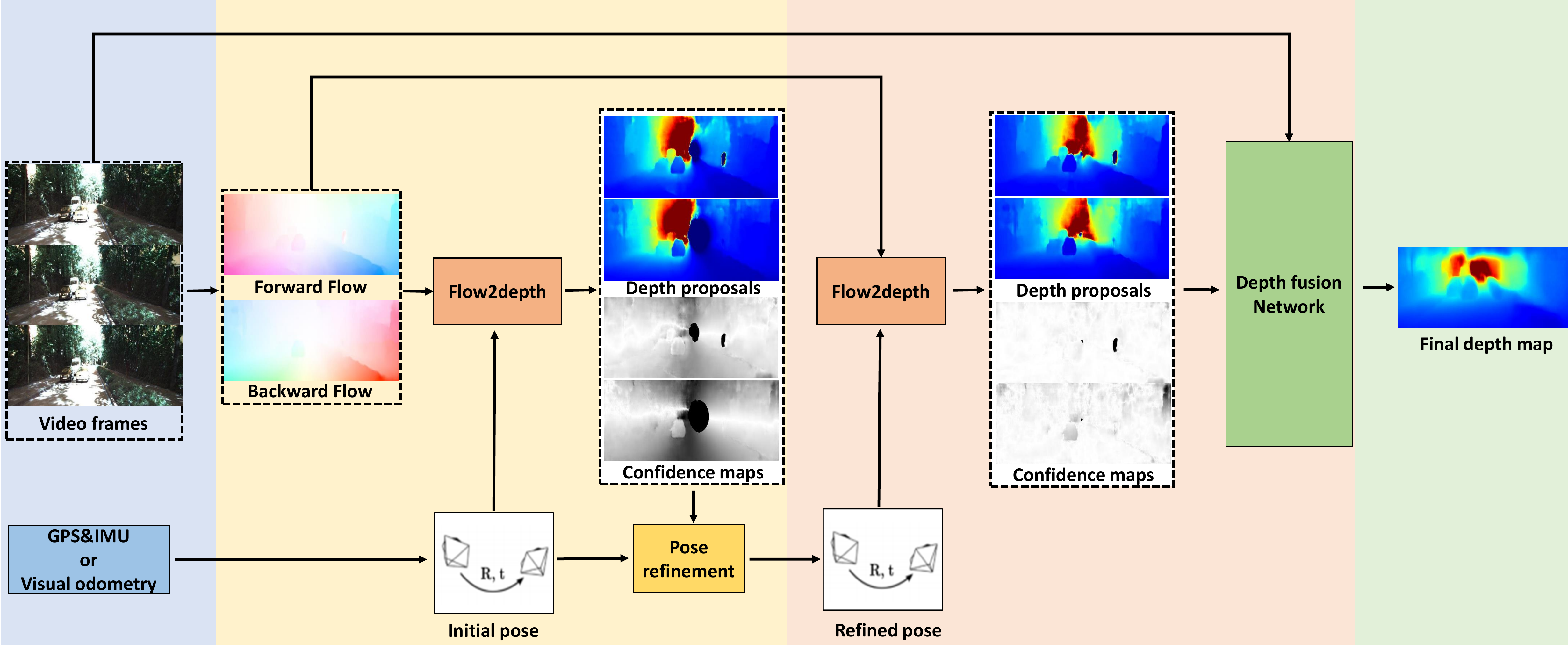}
\vspace{2mm}
\caption{The architecture of our overall framework. 
First, we estimate the optical flow from the video frames and obtain initial camera poses from GPS and IMU or applying odometry algorithms. Second, the initial camera poses are refined by maximizing the sum of confidence map in pose refinement module. Third, generating depth proposals and confidence maps with refined camera poses through the flow-to-depth layer. Finally, we obtain the final depth map by a depth fusion network that fuses the given depth proposals, confidence maps and target frame.
}
\label{fig:framework}
\end{figure*}

\section{INTRODUCTION}
Accurate dense depth estimation from a monocular video stream can be a backbone algorithm for autonomous robots and mobile devices. For autonomous ground or aerial vehicles, depth estimation from a monocular video can provide additional information for navigation and obstacle avoidance. A mobile device with a low-cost monocular camera can enable tremendous augmented reality applications without the need for dedicated depth sensors.


A line of research work on monocular depth estimation has been dedicated to single image depth estimation~\cite{SaxenaCN05, EigenF15, EgienPF14, FuCVPR18-DORN, LainaRBTN16}. However, single image depth estimation methods heavily rely on image priors learned from data, which might not generalize well to unseen scenes.
Since it is difficult to obtain extremely accurate depth maps from single image, some researchers focus on depth from video by utilizing multiple video frames~\cite{ummenhofer2017demon, YinShi2018, Ranftl2016, Zhou2017, deepv2d, Liu_2019_CVPR}. These approaches usually directly regress depth from deep features aggregated from multiple frames~\cite{Zhou2017} or cost volumes constructed by a plane-sweep algorithm~\cite{Liu_2019_CVPR}. Some methods use optical flow as part of the input to their network~\cite{ummenhofer2017demon} or as one auxiliary task~\cite{YinShi2018}. Different from these methods, our approach capitalizes on state-of-the-art optical flow methods to refine camera poses and generate depth proposals to improve the final depth estimation with a novel flow-to-depth layer. This flow-to-depth layer is built upon solving the classical triangulation problem for 3D depth estimation, and has the potential to generalize well to unseen environments.

One critical design in our model is a differentiable flow-to-depth layer that solves an epipolar geometry optimization problem. The flow-to-depth layer takes optical flow and camera poses as input and produces depth proposals. We show that our flow-to-depth layer does not only produce geometrically reliable depth maps (proposals) and the confidence of the depth but also helps refine the camera poses between video frames. At the end of our model, we have a fusion network that takes depth proposals and their confidence maps inferred from adjacent frames to produce the final depth maps. Note that the fusion network can additionally take the depth proposals generated by other depth estimation methods. For optical flow, we utilize the state-of-the-art optical flow methods that have gained significant progress~\cite{Sun2018PWC-Net}. To obtain the initial camera pose, we can use sensors such as IMU and GPS or apply odometry algorithms~\cite{Engel2018}.

We conduct extensive experiments on the KITTI~\cite{Geiger2012}, ScanNet~\cite{dai2017scannet}, and Waymo datasets~\cite{waymo_open_dataset}. The experiments show that our approach significantly outperforms state-of-the-art methods in depth estimation. Our controlled experiments indicate that the differentiable flow-to-depth layers in our model significantly improve the overall accuracy of video depth estimation by refining camera poses and generating depth proposals. To our surprise, our model trained on KITTI can generalize will to the unseen Waymo dataset while other methods do not. We believe the reason for the strong generalization capability of our model is that we solve for the depth proposals based on solving traditional triangulation problems rather than memorizing visual content. In summary, the main contributions of our work are as follows.
\begin{itemize}
    \item We present a novel framework with differentiable flow-to-depth layers for video depth estimation. The flow-to-depth layer refines camera poses and generates depth proposals by solving a triangulation problem between two video frames. 
    \item A depth fusion network can merge the depth proposals from the flow-depth-layer to produce the final depth maps. The depth fusion network can optionally take the depth maps generated by other methods to improve the performance further.
    \item We conduct thorough experiments on monocular depth estimation and show that our approach produces more accurate depth maps than contemporaneous methods do. Our model also demonstrates stronger generalization capability across datasets.
\end{itemize}

\section{Related Work}

In the literature, there is a large body of work on depth estimation from images. The settings can vary from single images, binocular stereo, temporal sequences to discrete multiple views. We briefly review them below.

\subsection{Single Image Depth}

Early work in this line can be traced back to Saxena et al.~\cite{HoiemEH05} and Hoiem et al.~\cite{SaxenaCN05}. The previous work learns to predict depth from single images using a discriminatively-trained Markov Random Field (MRF), while the later one classifies image pixels into different geometric regions, which can then be used to infer shapes. Most recently, with the success of deep learning, several works start to train deep convolutional neural networks to directly regress raw pixels to depth values~\cite{EgienPF14, EigenF15, LainaRBTN16,monodepth17, FuCVPR18-DORN}. Our work is fundamentally different from these approaches in that we take multiple images from a sequence to infer a more accurate structure from motions in regions where priors are less confident.

\subsection{Binocular Stereo Depth}

Depth estimation has been extensively exploited in the paired stereo setting, and the original problem is usually reformulated as a matching problem~\cite{ScharsteinS02}. Thus, traditional stereo approaches~\cite{Hirschmuller05, HosniRBRG13} often suffer from regions causing ambiguity in matching correspondence. Such regions include textureless areas, reflective surfaces, and repetitive patterns, to name a few. Most recently, deep learning has also shown its success in stereo matching~\cite{ZbontarL16, HanLJSB15}. The state-of-the-art approaches~\cite{KendallMDH17, ChangC18, YangZSDJ18, ChengWY18} usually construct a 3D cost volume and perform 3D convolutions on it.
Along with this direction, improvements have been made by pyramid~\cite{ChangC18}, semantic segmentation~\cite{YangZSDJ18}, learned affinity propagation~\cite{ChengWY18} and etc. The stereo pair setting usually generates an accurate depth and naturally adapts to dynamic scenes. However, the stereo pair rig is not ubiquitous in the real world, and thus it is less practical compared to the monocular setting.

\subsection{Depth from Video}
We predict depth from a monocular video sequence. Recent work includes~\cite{deepv2d, qifengcvpr2016depth, casser2018depth, ummenhofer2017demon}. Both ~\cite{qifengcvpr2016depth} and ~\cite{casser2018depth} explicitly models motion of moving objects. DeMoN~\cite{ummenhofer2017demon} proposes an architecture that alternates optical flow estimation with the estimation of camera motion and depth.
 Our work is closely related to DeepV2D ~\cite{deepv2d}, which leverages multi-view geometry to warp frames into a common viewpoint and constructs a cost volume from features of these frames to regress depth. DeepV2D also relates optical flow and depth but in a completely different way; it computes residual flow of the warped neighboring images, and utilize it to update camera poses. Unlike it, we estimate the rigid depth from flow via epipolar geometry. 

\subsection{Multi-view Reconstruction}
Multi-view stereo (MVS) reconstructs 3d shapes from a number of images, which is a core computer vision problem that has been studied for decades. Conventional MVS algorithms~\cite{HarlteyZ2006} perform 3D reconstruction by optimizing photometric consistency with handcrafted error functions to measure the similarity between patches. These algorithms lack the ability to handle poorly textured regions and reflective surfaces where photometric consistency is unreliable. 
Recent deep learning methods~\cite{yao2018mvsnet,huang2018deepmvs, im2018dpsnet} take a plane-sweep volume of deep features as input and produce depth maps for the reference images. Our method is fundamentally different from these cost volume-based methods in that we incorporate the multi-view geometry constraint via the flow-to-depth layer.  

\begin{figure}[t!]
\centering
\includegraphics[width=0.99\linewidth]{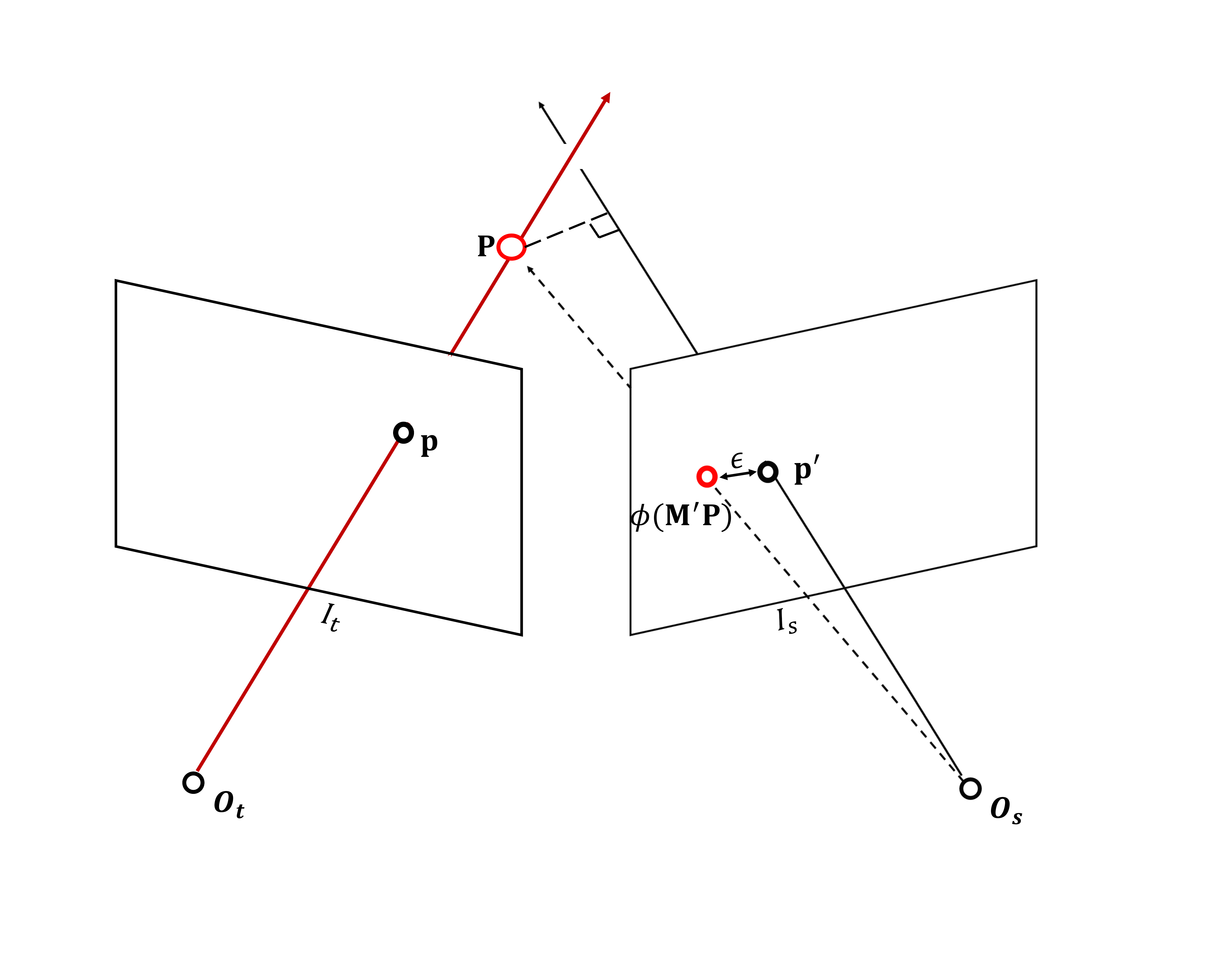}
\caption{Illustration of generating a depth proposal from optical flow. $\mathbf{p}$ and $\mathbf{p}'$ are corresponding points given by the optical flow. The objective of the flow-to-depth layer is to find an optimal  $\mathbf{P}$ that minimizes the reprojection error $\epsilon$.}
\label{fig:Flow2Depth}
\end{figure}

\section{Our Approach}
Given a sequence of frames $\left\{I_1, ..., I_N\right\}$ from a monocular video, our objective is to predict the depth map of every video frame by utilizing the frames around it. The input to our model includes the target frame $I_t$, its neighboring frames $\{I_s\}$ and the initial camera pose transformations $\{T_{t,s}\}$ between $I_t$ and $\{I_s\}$, which can be obtained from GPS and IMU, or by applying visual odometry algorithm~\cite{Nister2004}.

Fig.~\ref{fig:framework} illustrates the overall architecture of our proposed model, which consists of three critical components. First, the novel differentiable flow-to-depth layer. It takes optical flow and a camera pose as input and estimates rigid depth by triangulation in 3D. The layer produces both depth proposal map $D_{t, s}$ and confidence map $C_{t, s}$ for the target frame by optimizing an epipolar geometry least square problem. 

Second, the camera pose refinement module. The initial relative camera pose $T_{t,s}$ may not be highly accurate due to noisy sensor outputs from GPS and IMU, or imperfect visual odometry algorithms. Since flow-to-depth layer is differentiable, we can use it to backpropagate the gradients from the confidence map to the initial camera pose and refine the initial camera pose by maximizing the sum of confidence map. Our experimental results show that the pose refinement module significantly improves performance.

The last one is depth fusion network. It takes target frame, depth proposals and confidence maps to generate the final depth map $D$. The intuition behind such a depth fusion network is that, for regions with high confidence, the network can directly use the provided depth values; otherwise, the network will perform depth interpolation or inpainting. Note that we also provide the target frame as an additional input to the depth fusion network, which provides strong image priors for inpainting the regions with low confidence. We find that utilizing depth proposals along with their confidence maps greatly improves the depth estimation quality. 

\begin{figure}[t!]
\centering
\begin{tabular}{@{}c@{\hspace{1mm}}c@{\hspace{1mm}}c@{}}
&Scene 1&Scene 2\\
\rotatebox{90}{\small \hspace{3.2mm} $I_t$}&
\includegraphics[width=0.45\linewidth]{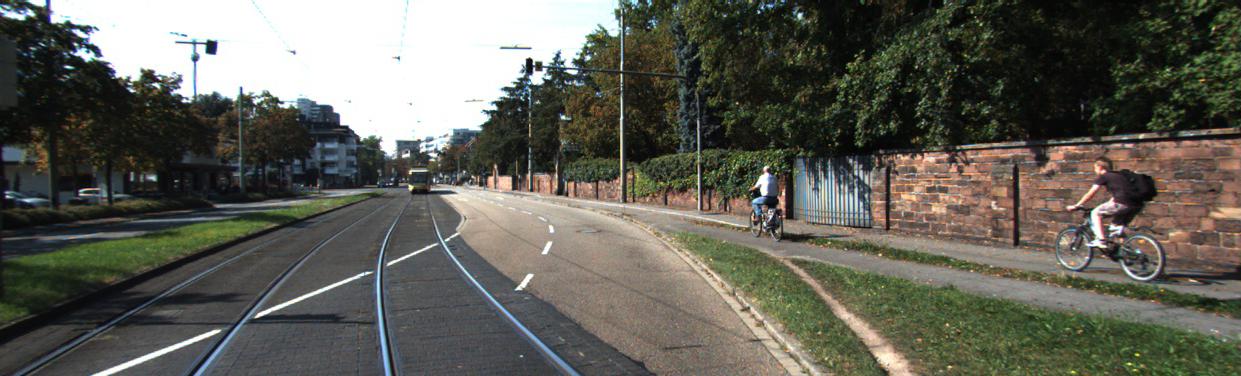}&
\includegraphics[width=0.45\linewidth]{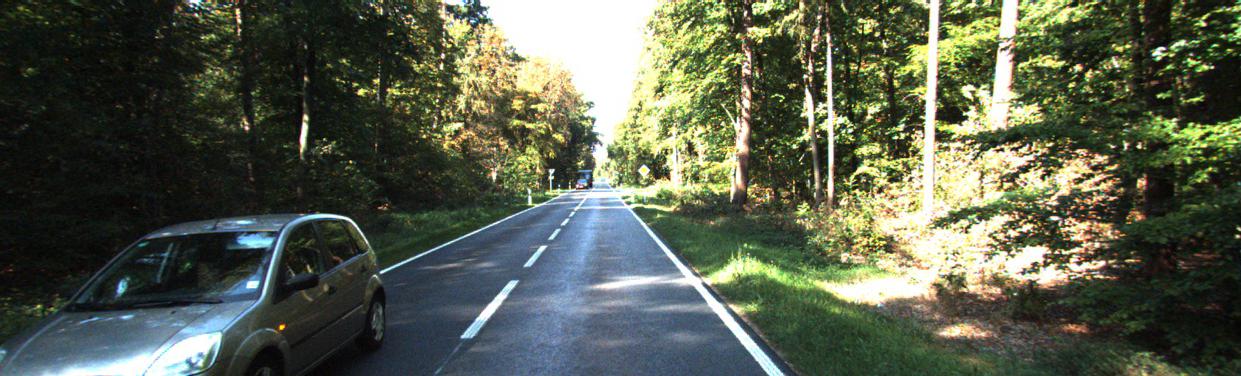}\\

\rotatebox{90}{\small \hspace{2.2mm} $C_{t,s}$}&
\includegraphics[width=0.45\linewidth]{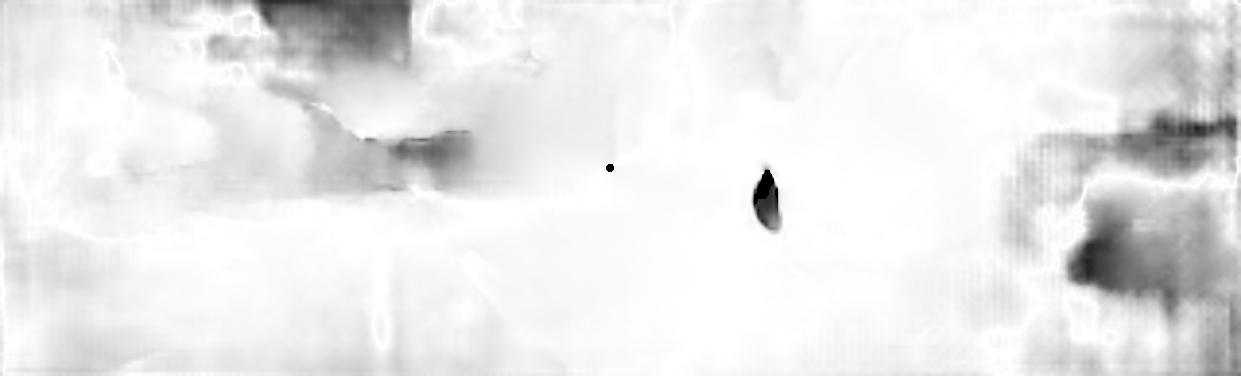}&
\includegraphics[width=0.45\linewidth]{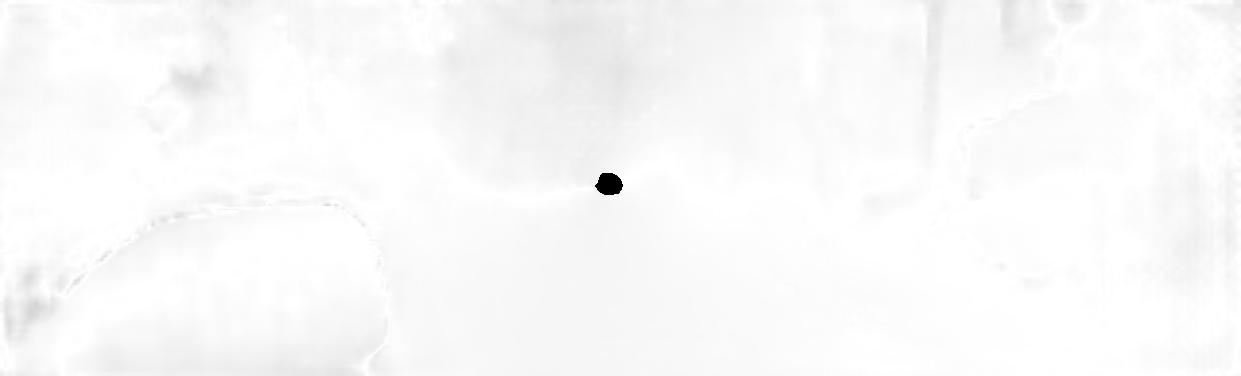}\\

\rotatebox{90}{\small \hspace{2.2mm} $D_{t,s}$}&
\includegraphics[width=0.45\linewidth]{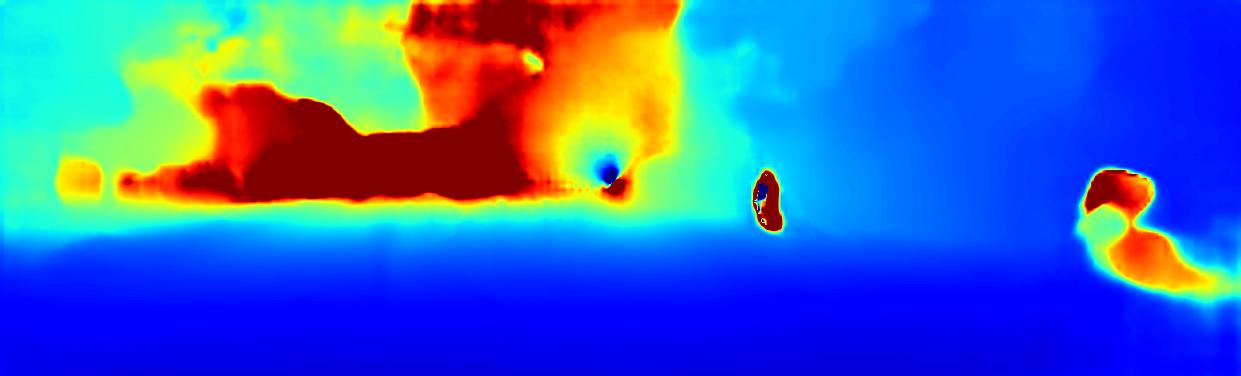}&
\includegraphics[width=0.45\linewidth]{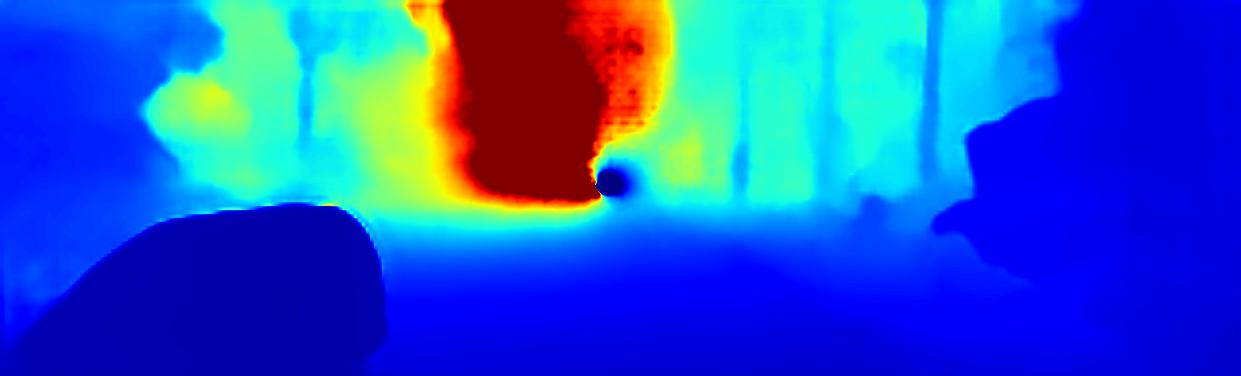}\\

\end{tabular}
\vspace{2mm}
\caption{Confidence maps (the second row) and depth proposals (the third row) generated by the flow-to-depth layer on the KITTI dataset. For the confidence maps, darker areas indicate lower confidence. For depth proposal, blue areas indicate small depth values.}
\label{fig:Ablation-div}
\end{figure}

\subsection{Flow-to-depth Layers}
Parallax can appear between two adjacent video frames because of camera motion. We utilize this parallax to improve monocular depth estimation by introducing a differentiable flow-to-depth layer.

\paragraph{Depth proposals}
Consider the depth estimation problem for a target frame $I_t$, given a nearby source frame $I_s$, we leverage optical flow and relative camera pose between $I_t,I_s$ to generate a depth proposal $D_{t,s}$. Fig.~\ref{fig:Flow2Depth} illustrates configuration of our problem. Using homogeneous coordinate, assume a 3D point $\mathbf{P}=[x, y, d, 1]^T$ with its corresponding pixels in $I_t$ and $I_s$ being $\mathbf{p}=[u,v,1]^{T},\mathbf{p}'=[u',v',1]^{T}$, respectively. Given $\mathbf{p}$ and $\mathbf{p}'$, we solve for an optimal $\mathbf{P}$ that minimizes the reprojection error. Let the world coordinate system be the camera coordinate system of $I_t$. Suppose $\mathbf{M^{'}}$ is the camera matrix for $I_s$, and $\mathbf{K}$ is the intrinsic matrix for $I_t$. In the following, we use numbers in subscript to slice vectors and matrices and use comma to separate dimensions. Then we have
\begin{align}
\mathbf{p}=\mathbf{K}\mathbf{P_{1:3}},  \mathbf{p'}=\mathbf{M'}\mathbf{P}.
\end{align}
Our reprojection error is formulated as:
\begin{align}
   \epsilon(d) =\| \phi( \mathbf{M}'\begin{bmatrix}d\mathbf{K}^{-1}\mathbf{p}\\ 1\end{bmatrix})-\mathbf{p}'\|,
\end{align}
where $d$ is the depth of $\mathbf{P}$, and $\phi(\mathbf{x})=\frac{\mathbf{x}}{\mathbf{x}_3}, \mathbf{x}\in \mathcal{R}^3$.
For notation convenience, denote $\mathbf{a}=\mathbf{M'}_{1:3,1:3}\mathbf{K}^{-1}\mathbf{p},\mathbf{b}=\mathbf{M}'_{1:3, 4}$. Then the optimal $d^*$ minimizing $\epsilon(d)$ can be computed in closed form:
\begin{align}
    d^*= \arg\min_d {\| \phi(d \mathbf{a}+\mathbf{b})-\mathbf{p}'\|}=\frac{1}{\mathbf{m}^T\mathbf{m}}\mathbf{m}^T\mathbf{n},
    \label{eq:optimal_depth}
\end{align}
where $
    \mathbf{m}=\mathbf{a}_{1:2}-\mathbf{a}_3 \mathbf{p}'_{1:2}, 
    \mathbf{n}=\mathbf{b}_3\mathbf{p}'_{1:2}-\mathbf{b}_{1:2}.
$

We use optical flow algorithms, e.g., PWC-Net~\cite{Sun2018PWC-Net}, to find dense pixel-wise correspondences between $I_t, I_s$, then solve for the optimal depth at each pixel location with Equation \eqref{eq:optimal_depth}. Since this procedure is differentiable, we implement it as a flow-to-depth layer to enable end-to-end training.

\paragraph{Confidence maps}
The reprojection error $\epsilon$ can serve as a confidence measure for the computed depth: the larger the reprojection error is, the more likely the depth is prone to error. We obtain a confidence map $C_{t,s}$ by converting $\epsilon$ into confidence using $\exp{(-\frac{\epsilon}{\sigma})}$, where $\sigma$ is a normalization constant. We set $\sigma=20$ in experiments. Moreover, if the computed $d$ is negative, we set its confidence to zero. Fig.~\ref{fig:Ablation-div} shows our depth proposals and the corresponding confidence maps.

\subsection{Camera Pose Refinement}

The quality of our depth proposals highly depend on the camera poses. In practice, we can obtain an initial camera pose from sensors such as GPS and IMU, but the initial camera pose is not highly accurate due to sensor noise. To improve the accuracy of camera pose, we utilize our flow-to-depth layer to refine the camera pose. 

We can refine the camera pose by building the relationship between the camera pose and the confidence map through the differentiable flow-to-depth layer. Typically, a good camera pose should lead to a large confidence map. We define a maximizing objective function to optimize the camera pose $T_{t,s}$:
\begin{equation}
\label{eqn:pose}
    L(T_{t,s})= \sum_{p \in \mathcal{S}}{C_{t,s}(p)},
\end{equation}
where $ S $ is the set of pixels with positive depth in the depth proposal. We exclude those pixels with negative depths because we do not use the negative depth at all. The objective is designed to maximize the sum of the confidence on each pixel with positive depth.

To minimize the objective function in \eqref{eqn:pose}, we use L-BFGS-B optimizer~\cite{zhu1997algorithm}, and set bounds $[-\pi,\pi]$ for rotation in $M$.  Note that we are able to compute the gradients on the camera pose because the flow-to-depth layer is differentiable. We evaluate the performance with and without the pose refinement, and experiments show that the refinement can significantly improve the depth estimation.

\subsection{Depth Fusion}
For each pair of the target frame and the source frame, we can generate a depth proposal and a confidence map. Then we can make use of depth proposals and confidence maps to produce a high-quality final depth map. Our depth fusion network is designed to merge them and perform refinement as needed. Compared to single image depth estimation methods, our approach has the benefits that the model can take advantage of the depth proposals and their confidence maps for better depth estimation.


As shown in Fig.~\ref{fig:framework}, we concatenate the target frame $I_{t}$, depth proposals $D_{t,s}$, confidence maps $C_{t,s}$ as input to the depth fusion network. The output of the depth fusion network is the final depth map $D$. Besides the depth proposals and confidence maps computed by our flow-to-depth layer, our depth fusion network can also take the depth proposals generated by other methods to improve the estimation accuracy. In addition, we find that fusing the depth maps generated by other methods to further improve the performance.
We train our depth fusion network with provided ground-truth depth maps in a supervised fashion.


\paragraph{Loss function}
 Our depth loss is defined over each pixel $p$ with ground-truth depth:
\begin{equation}
     L_{depth}=\sum_{p}||\log D_p-\log\hat{D}_p||,
\end{equation}
where $\hat{D}$ is the ground-truth depth map. We define the depth loss in the log domain rather than the linear domain because this can avoid distant pixels dominating the loss. 

We also use a smoothness loss by imposing smoothness regularization on the output disparity map (inverse of the depth map). The smoothness loss is defined as
\begin{equation}
    L_{smooth} = \sum_p{\nabla^{2} \frac{1}{D_p}},
\end{equation}
where $\nabla^{2}$ is the Laplacian operator.

The total loss for the depth fusion network is
\begin{equation}
    L_{fusion} = \lambda_d L_{depth} + \lambda_s L_{smooth},
\end{equation}
where $\lambda_{d}=1$ and $\lambda_{s}=0.5$.

\paragraph{Network architecture}
Our depth fusion network adopts the single view depth network in SfMLearner~\cite{Zhou2017}. It is an encoder-decoder architecture with skip connections and multi-scale prediction.

\vspace{2mm}
\begin{table*}[t]
\renewcommand\arraystretch{1.7}
\begin{center}
\caption{Quantitative evaluation of KITTI dataset}
\begin{tabular}{@{}l@{\hspace{5mm}}l@{\hspace{5mm}}c@{\hspace{4mm}}c@{\hspace{4mm}}c@{\hspace{4mm}}c@{\hspace{4mm}}c@{\hspace{4mm}}c@{\hspace{4mm}}c@{\hspace{4mm}}c@{\hspace{4mm}}c@{\hspace{4mm}}c@{}}
\hline
Method & Type &split& abs rel $\downarrow$ & sq rel $\downarrow$ & rms $\downarrow$ & log rms $\downarrow$ & irmse $\downarrow$& SIlog $\downarrow$ & $\delta_1 \uparrow$ & $\delta_2 \uparrow $ & $\delta_3 \uparrow$ \\
\hline
Eigen et al.~\cite{EgienPF14} coarse &supervised&Eigen& 0.194 & 1.531 & 7.216 & 0.273 & - & - & 0.679 & 0.897 & 0.967\\
Eigen et al.~\cite{EgienPF14} fine&supervised&Eigen&0.190 & 1.515 & 7.156 & 0.270 & - & - & 0.692 & 0.899 & 0.967\\
GeoNet~\cite{YinShi2018} &unsupervised&Eigen&0.155 & 1.297 & 5.857 & 0.233 & 0.018& 0.229 & 0.793 & 0.931 & 0.973\\
Godard et al.~\cite{monodepth17} &unsupervised+stereo&Eigen&0.150& 1.329 & 5.806 & 0.231 & 0.019 & 0.227 & 0.810 & 0.933 & 0.971\\
Kuznietsov et al.~\cite{Kuznietsov_2017_CVPR_semi_depth} &semi-supervised+stereo&Eigen&0.110 & 0.708 & 4.312 & 0.172 & 0.014 & 0.169& 0.878 & 0.964 & 0.987\\
DORN~\cite{FuCVPR18-DORN} &supervised&Eigen&0.102& 0.592 & 3.837 & 0.162 & 0.015 & 0.158& 0.898 & 0.967 & 0.986\\
Ours&supervised+video&Eigen&\textbf{0.081}&\textbf{0.488}&\textbf{3.651}&\textbf{0.146} &\textbf{0.012} &\textbf{0.144}&\textbf{0.912}&\textbf{0.970}&\textbf{0.988}\\
\hline
NeuralRGBD~\cite{Liu_2019_CVPR}&supervised+video&Uhrig&0.105 & 0.532 & 3.299 & 0.150 & 0.013 & 0.144 & 0.887 & 0.972 & 0.990\\ 
Ours&supervised+video&Uhrig&\textbf{0.071}&\textbf{0.338}&\textbf{2.537}&\textbf{0.116} &\textbf{0.010}&\textbf{0.112}&\textbf{0.938}&\textbf{0.979}&\textbf{0.992}\\
\hline
\end{tabular}
\end{center}
\label{table:KITTI dataset comparison}
\end{table*}

\begin{figure*}[t]
\centering
\begin{tabular}{@{}c@{\hspace{0.7mm}}c@{\hspace{0.7mm}}c@{\hspace{0.7mm}}c@{}}
&Scene 1 & Scene 2 & Scene 3\\
\rotatebox{90}{\small \hspace{3.5mm} $I_t$}&
\includegraphics[width=0.32\linewidth,height=0.07\linewidth]{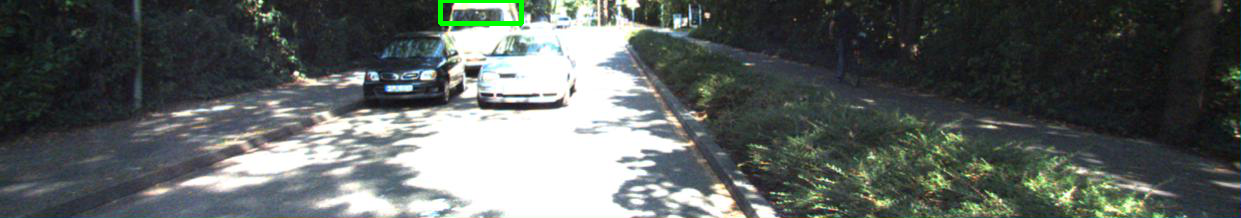}&
\includegraphics[width=0.32\linewidth,height=0.07\linewidth]{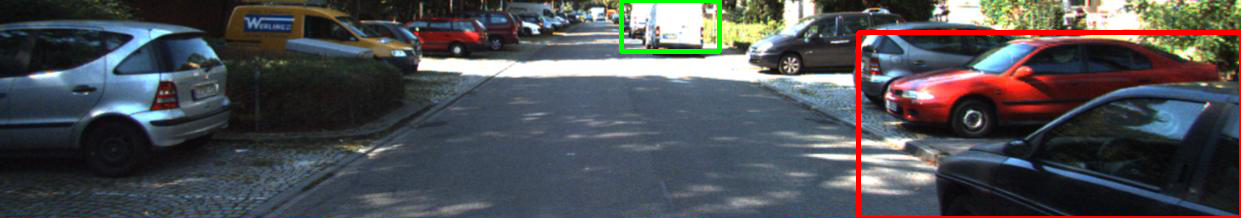}&
\includegraphics[width=0.32\linewidth,height=0.07\linewidth]{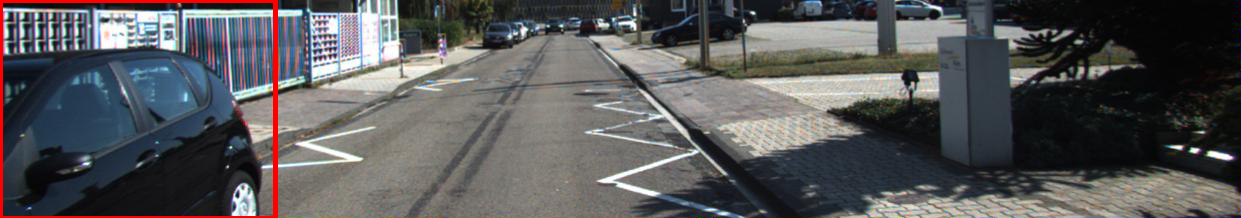}\\
\rotatebox{90}{\small \hspace{3.5mm}  GT}&
\includegraphics[width=0.32\linewidth,height=0.07\linewidth]{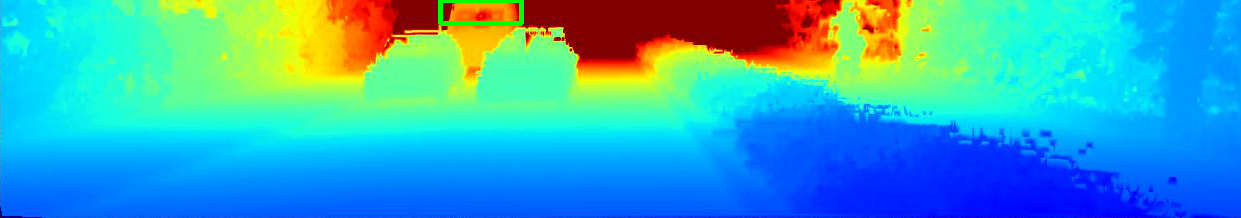}&
\includegraphics[width=0.32\linewidth,height=0.07\linewidth]{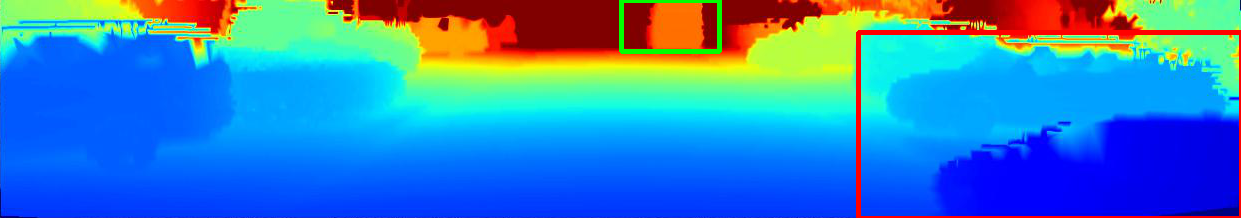}&
\includegraphics[width=0.32\linewidth,height=0.07\linewidth]{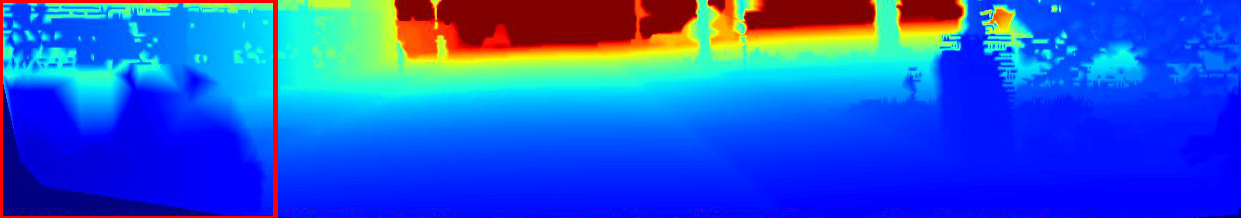}\\
\rotatebox{90}{\small \hspace{2.5mm} ~\cite{FuCVPR18-DORN}}&
\includegraphics[width=0.32\linewidth,height=0.07\linewidth]{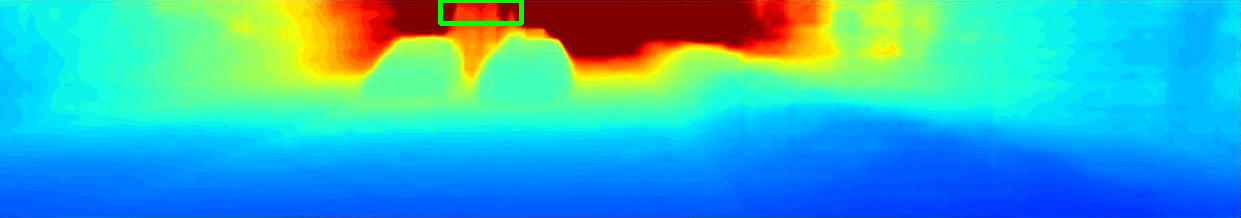}&
\includegraphics[width=0.32\linewidth,height=0.07\linewidth]{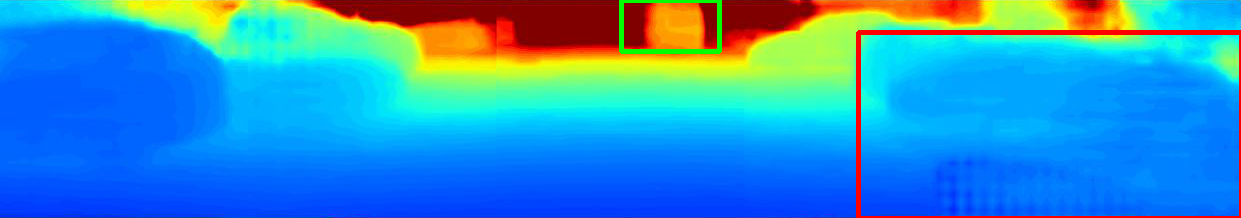}&
\includegraphics[width=0.32\linewidth,height=0.07\linewidth]{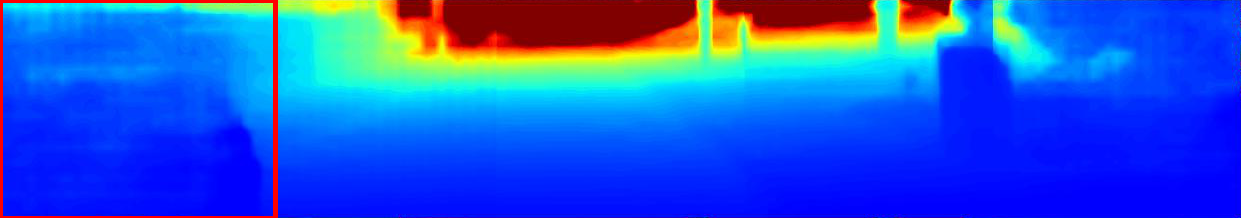}\\

\rotatebox{90}{\small \hspace{1.8mm} ~\cite{Liu_2019_CVPR}}&
\includegraphics[width=0.32\linewidth,height=0.07\linewidth]{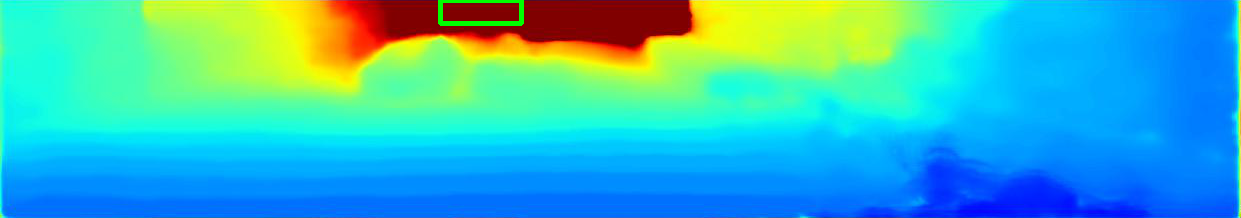}&
\includegraphics[width=0.32\linewidth,height=0.07\linewidth]{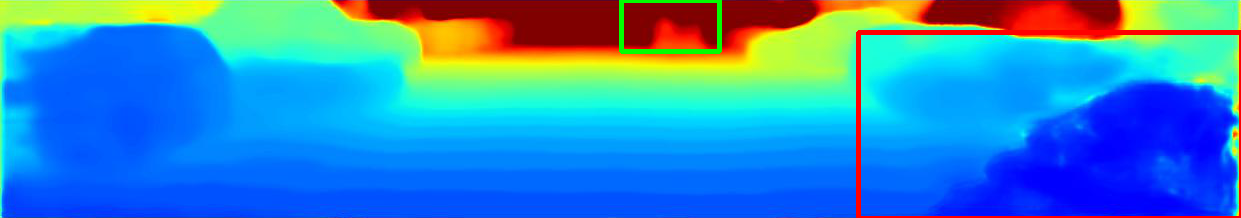}&
\includegraphics[width=0.32\linewidth,height=0.07\linewidth]{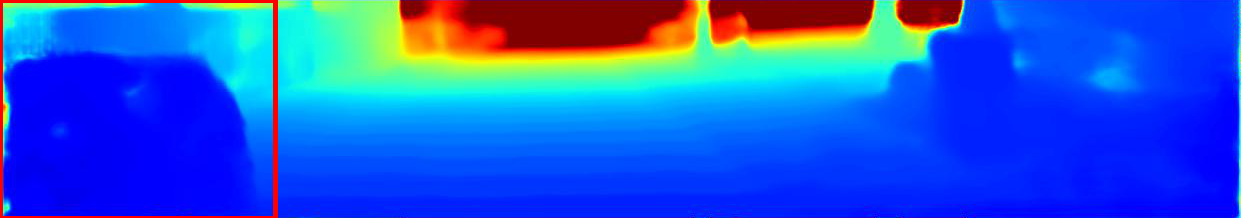}\\

\rotatebox{90}{\small \hspace{2mm}  Ours}&
\includegraphics[width=0.32\linewidth,height=0.07\linewidth]{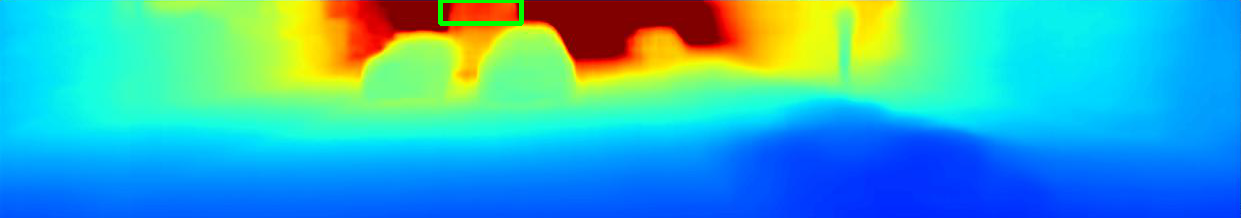}&
\includegraphics[width=0.32\linewidth,height=0.07\linewidth]{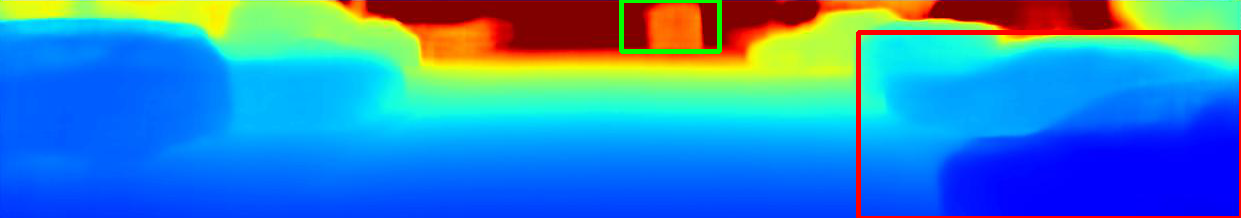}&
\includegraphics[width=0.32\linewidth,height=0.07\linewidth]{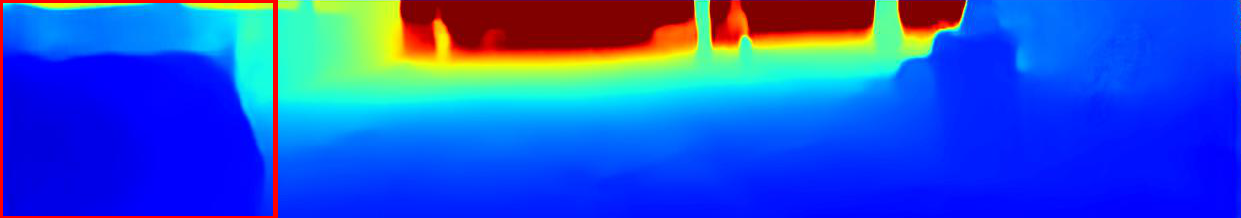}\\
\end{tabular}

\caption{Qualitative comparisons of DORN~\cite{FuCVPR18-DORN}, NeuralRGBD~\cite{Liu_2019_CVPR}, and ours on the KITTI dataset. The ground-truth depth map is interpolated from sparse measurements for the visualization purpose.}
\label{fig:KITTI}
\end{figure*}

\section{Experiments}
\subsection{Implementation}

For the depth estimation of the target frame $I_t$, we use $I_{t-k_{1}}$ and  $I_{t+k_{2}}$ as the source frames. Since depth proposals have poor results when the camera translation between $I_t$ and $I_s$ (defined as $\|O_{s}-O_t\|$) is too small, we search for the smallest $k_{1}$ that satisfies $\|O_{t-k_{1}}-O_t\|>T$ where $T$ is a threshold. For KITTI and Waymo dataset, $T$ is 80cm. For ScanNet dataset, it is 12cm. We perform the similar search for $k_{2}$.

For training the model, we use the Adam optimizer~\cite{Kingma2015} with the learning rate of $0.0001$, batch size of $4$, $\beta_1=0.9$, and $\beta_2=0.999$. We use full-resolution video frames and ground-truth depth maps during training and evaluation.

\subsection{Datasets} We conduct experiments on three datasets: the KITTI dataset~\cite{Geiger2012}, the ScanNet dataset~\cite{dai2017scannet}, and the Waymo dataset~\cite{waymo_open_dataset}. 

The KITTI dataset contains outdoor images with depth maps projected from point clouds and also provides camera pose calculated from GPS and IMU. To compare with different previous works, we train our method in two different splits. One is the Eigen split proposed by Eigen et al.~\cite{EgienPF14}, in which the train set contains 33 video scenes,  the test set consists of 697 images extracted from 28 video scenes, and ground-truth depth maps projected from single-frame point clouds. Another one is the Uhrig split~\cite{Uhrig2017THREEDV} that came with the KITTI single image depth prediction benchmark. It has 138 training video scenes and 13 validation scenes. We randomly sample 50 images from every video scene in the validation set and get a test set consists of 650 images. Meanwhile, this split provides denser ground-truth depth maps, which are accumulated by 11 consecutive frames point clouds. Since different video sequences in KITTI may have different image sizes, we resize all the images to $376\times 1241$.
 
The ScanNet dataset is an RGB-D video dataset containing 2.5 million views in more than 1500 scans, annotated with 3D camera poses, surface reconstructions, depth maps, and instance-level semantic segmentations. For the train set and test set, we follow the instruction of the Robust Vision Challenge 2018 Workshop at CVPR 2018. 

The Waymo Open Dataset is a recently released autonomous driving dataset. It contains LiDAR and camera data from 1,000 video segments, splited into training set and validation set. We randomly sample 5 images from every daytime validation video segment and obtain a total of 784 test images to do cross dataset experiment.

\subsection{Baselines}  
In the KITTI Eigen split, we compare our method with several state-of-the-art depth estimation approaches: DORN~\cite{FuCVPR18-DORN}, Kuznietsov et al.~\cite{Kuznietsov_2017_CVPR_semi_depth}, Godard et al.~\cite{monodepth17}, GeoNet~\cite{YinShi2018}, and Eigen et al.~\cite{EgienPF14}. 

In the KITTI Uhrig split, we compare our method against state-of-the-art video depth estimation approach: NeuralRGBD~\cite{Liu_2019_CVPR}. We re-train NeuralRGBD~\cite{Liu_2019_CVPR} in the Eigen split, but its results are poor. To have a fair comparison, we also train our method in the Uhrig split and compare it with the results by the pre-trained model of NeuralRGBD~\cite{Liu_2019_CVPR}.

In the ScanNet and Waymo datasets, We carefully select two deep learning based methods for comparisons. For supervised depth estimation approaches, we choose DORN~\cite{FuCVPR18-DORN}, which is state of the art. For video depth estimation methods, we choose NeuralRGBD~\cite{Liu_2019_CVPR} that is highly related to our work.

\vspace{2mm}
\begin{table*}
\renewcommand\arraystretch{1.5}
\begin{center}
\caption{Quantitative evaluation of ScanNet dataset}
\begin{tabular}{@{}l@{\hspace{7mm}}l@{\hspace{7mm}}c@{\hspace{7mm}}c@{\hspace{7mm}}c@{\hspace{7mm}}c@{\hspace{7mm}}c@{\hspace{7mm}}c@{\hspace{7mm}}c@{\hspace{7mm}}c@{\hspace{7mm}}c@{}}
\hline
Method & Type & abs rel $\downarrow$ & sq rel $\downarrow$ & rms $\downarrow$ & log rms $\downarrow$ & irmse $\downarrow$& SIlog $\downarrow$ & $\delta_1 \uparrow$ & $\delta_2 \uparrow $ & $\delta_3 \uparrow$ \\
\hline
DORN~\cite{FuCVPR18-DORN}&supervised&0.096&0.033&0.217&0.127&0.099&0.120&0.907&0.981&\textbf{0.996}\\
NeuralRGBD~\cite{Liu_2019_CVPR}
&supervised&0.097&0.050&0.249&0.132&0.093&0.126&0.906&0.975&0.993\\
\hline
Ours&supervised&\textbf{0.076}&\textbf{0.029}&\textbf{0.199}&\textbf{0.108}&\textbf{0.077}&\textbf{0.103}&\textbf{0.933}&\textbf{0.984}&\textbf{0.996} \\
\hline
\end{tabular}
\end{center}
\label{table:ScanNet comparison}
\end{table*}

\begin{table*}
\renewcommand\arraystretch{1.5}
\begin{center}
\caption{Quantitative evaluation of Waymo dataset}
\begin{tabular}{@{}l@{\hspace{7mm}}l@{\hspace{7mm}}c@{\hspace{7mm}}c@{\hspace{7mm}}c@{\hspace{7mm}}c@{\hspace{7mm}}c@{\hspace{7mm}}c@{\hspace{7mm}}c@{\hspace{7mm}}c@{\hspace{7mm}}c@{}}
\hline
Method & Type& abs rel $\downarrow$ & sq rel $\downarrow$ & rms $\downarrow$ & log rms $\downarrow$ & irmse $\downarrow$& SIlog $\downarrow$ & $\delta_1 \uparrow$ & $\delta_2 \uparrow $ & $\delta_3 \uparrow$ \\
\hline
SfMLearner~\cite{Zhou2017}
&unsupervised&0.514&7.878&16.029&0.587&0.031&0.579&0.256&0.487&0.703\\
DORN~\cite{FuCVPR18-DORN}
&cross dataset&0.389&5.056&12.432&0.451&0.024&0.442&0.353&0.660&0.867 \\
NeuralRGBD~\cite{Liu_2019_CVPR}
&cross dataset&0.177&2.646&9.891&0.402&0.072&0.396&0.790&0.921&0.958 \\
\hline
Ours&cross dataset&\textbf{0.150}&\textbf{1.691}&\textbf{6.773}&\textbf{0.222}&\textbf{0.013}&\textbf{0.211}&\textbf{0.804}&\textbf{0.924}&\textbf{0.966}\\
\hline
\end{tabular}
\end{center}
\label{table:waymo comparison}
\end{table*}

\begin{figure*}
\centering
\begin{tabular}{@{}c@{\hspace{0.7mm}}c@{\hspace{0.7mm}}c@{\hspace{0.7mm}}c@{\hspace{0.7mm}}c@{\hspace{0.7mm}}c@{}}

&&&DORN~\cite{FuCVPR18-DORN}&NeuralRGBD~\cite{Liu_2019_CVPR}&Ours\\
\rotatebox{90}{\small  \hspace{12mm} $I_t$}&
\includegraphics[width=0.22\linewidth]{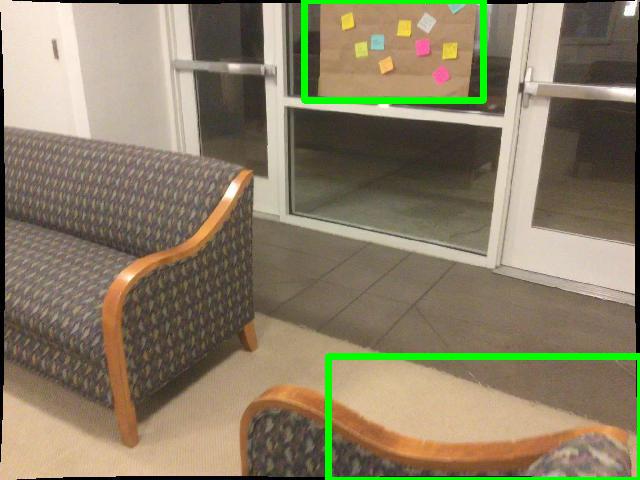}&
\rotatebox{90}{\small  \hspace{6mm} Depth map}&
\includegraphics[width=0.22\linewidth]{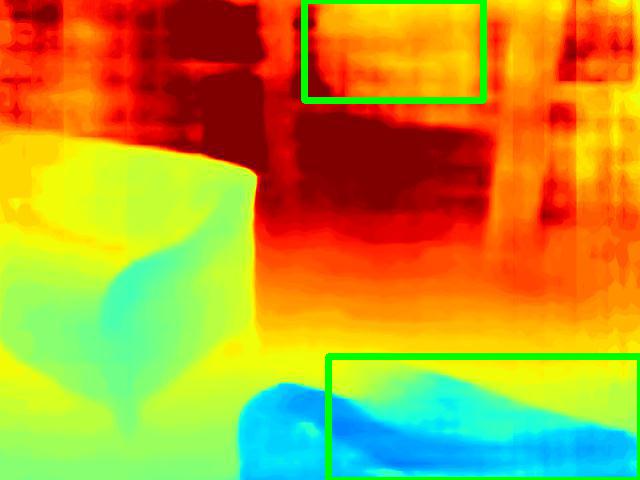}&
\includegraphics[width=0.22\linewidth]{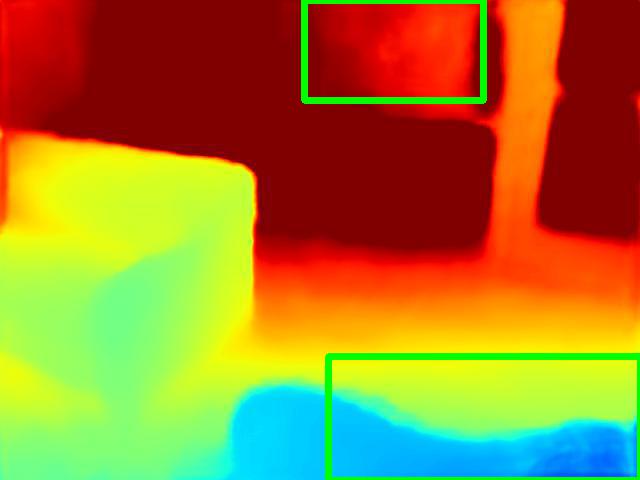}&
\includegraphics[width=0.22\linewidth]{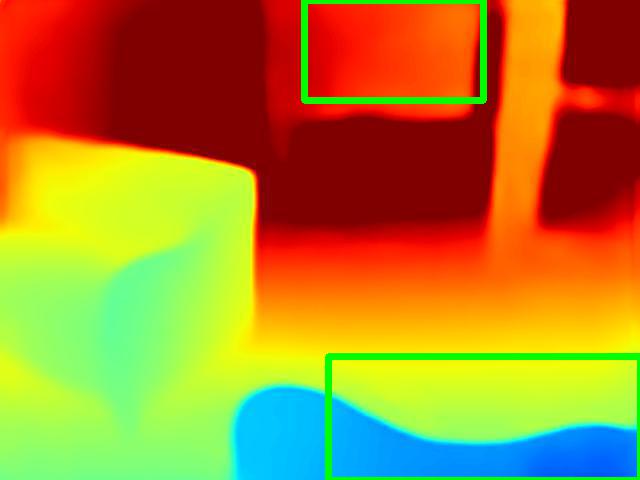}\\

\rotatebox{90}{\small  \hspace{12mm}  GT}&
\includegraphics[width=0.22\linewidth]{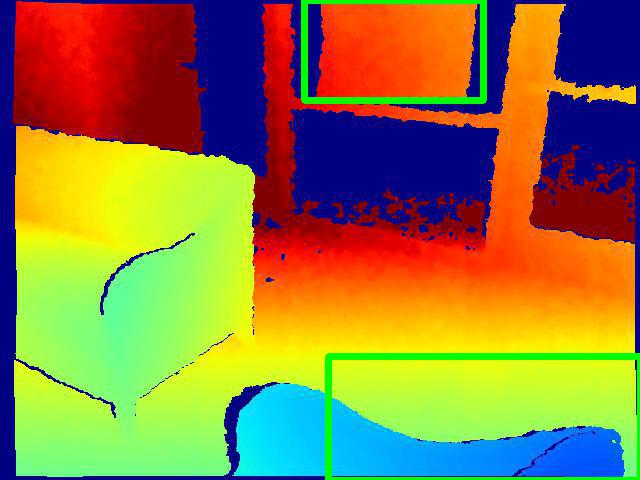}&
\rotatebox{90}{\small  \hspace{6mm} Error map}&
\includegraphics[width=0.22\linewidth]{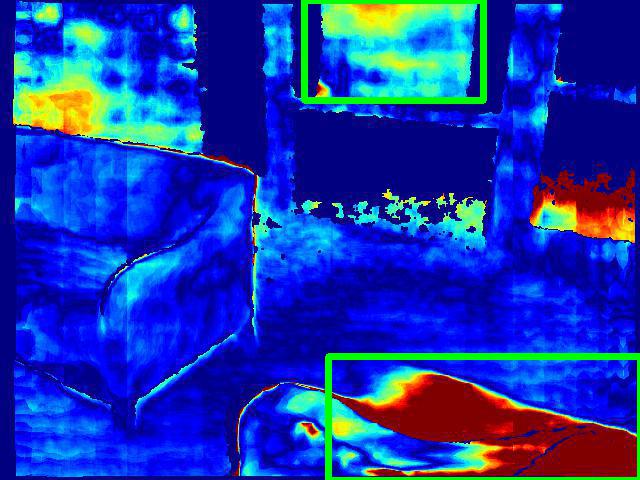}&

\includegraphics[width=0.22\linewidth]{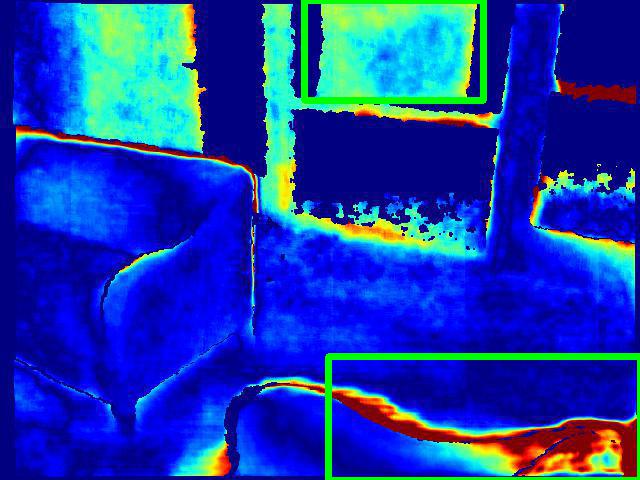}&

\includegraphics[width=0.23\linewidth]{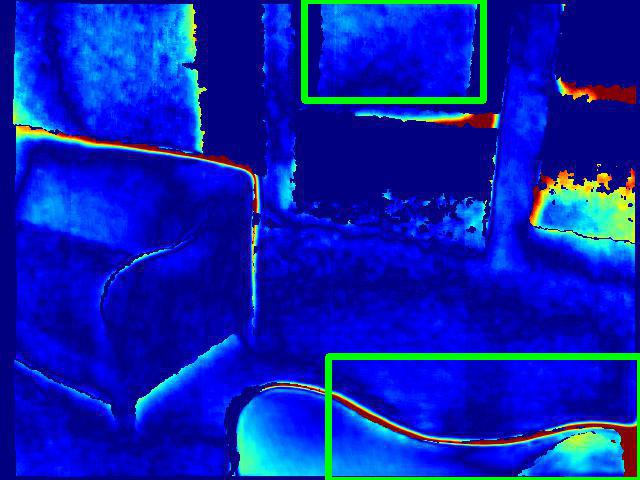}\\

\end{tabular}
\caption{Qualitative comparisons between DORN~\cite{FuCVPR18-DORN}, NeuralRGBD~\cite{Liu_2019_CVPR}, and ours on the ScanNet dataset. For the error maps, blue areas indicate low errors and red areas indicate high errors.}
\label{fig:scanNet results}
\end{figure*}

\begin{figure*}
\centering
\begin{tabular}{@{}c@{\hspace{0.7mm}}c@{\hspace{0.7mm}}c@{\hspace{0.7mm}}c@{}}



\rotatebox{90}{\small  \hspace{8mm}$I_t$}
\includegraphics[width=0.24\linewidth]{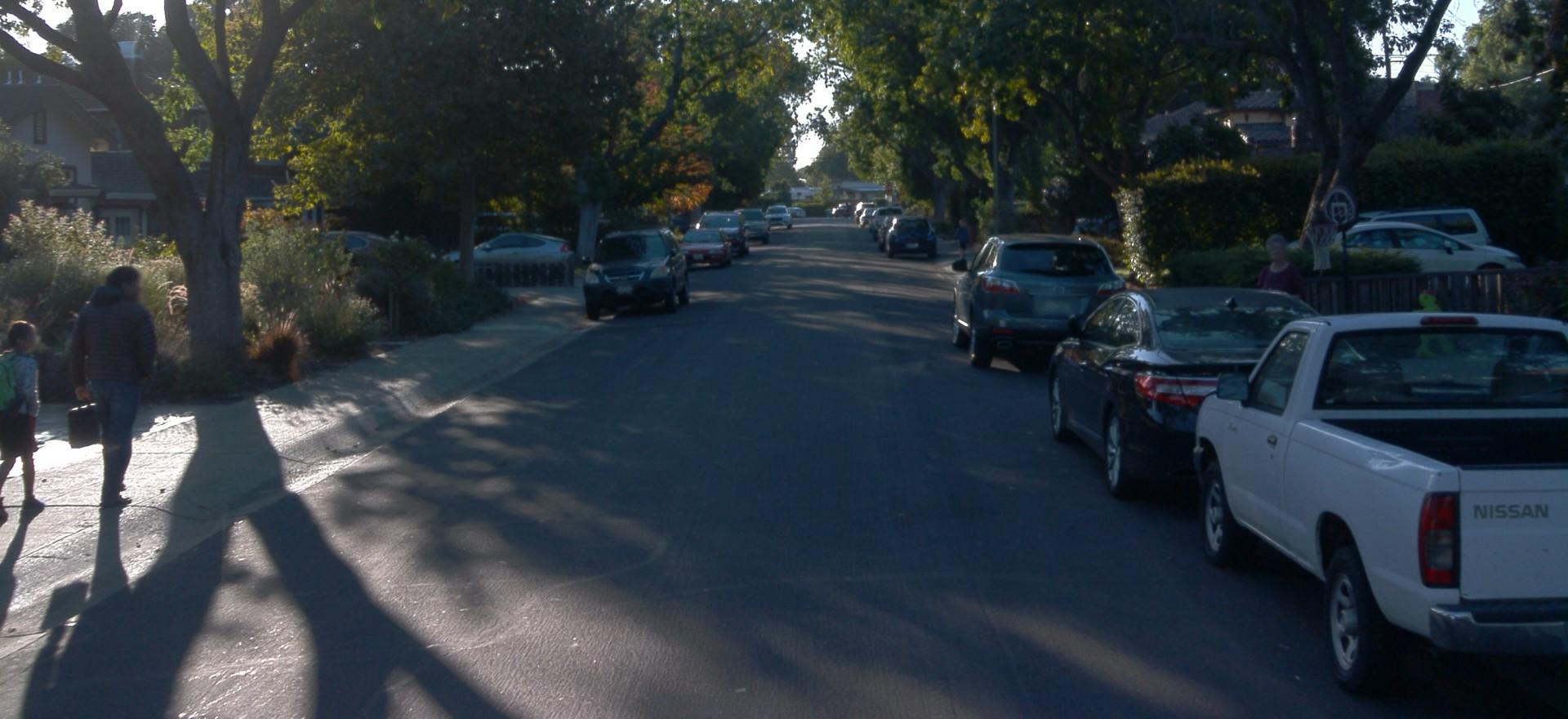}&
\includegraphics[width=0.24\linewidth]{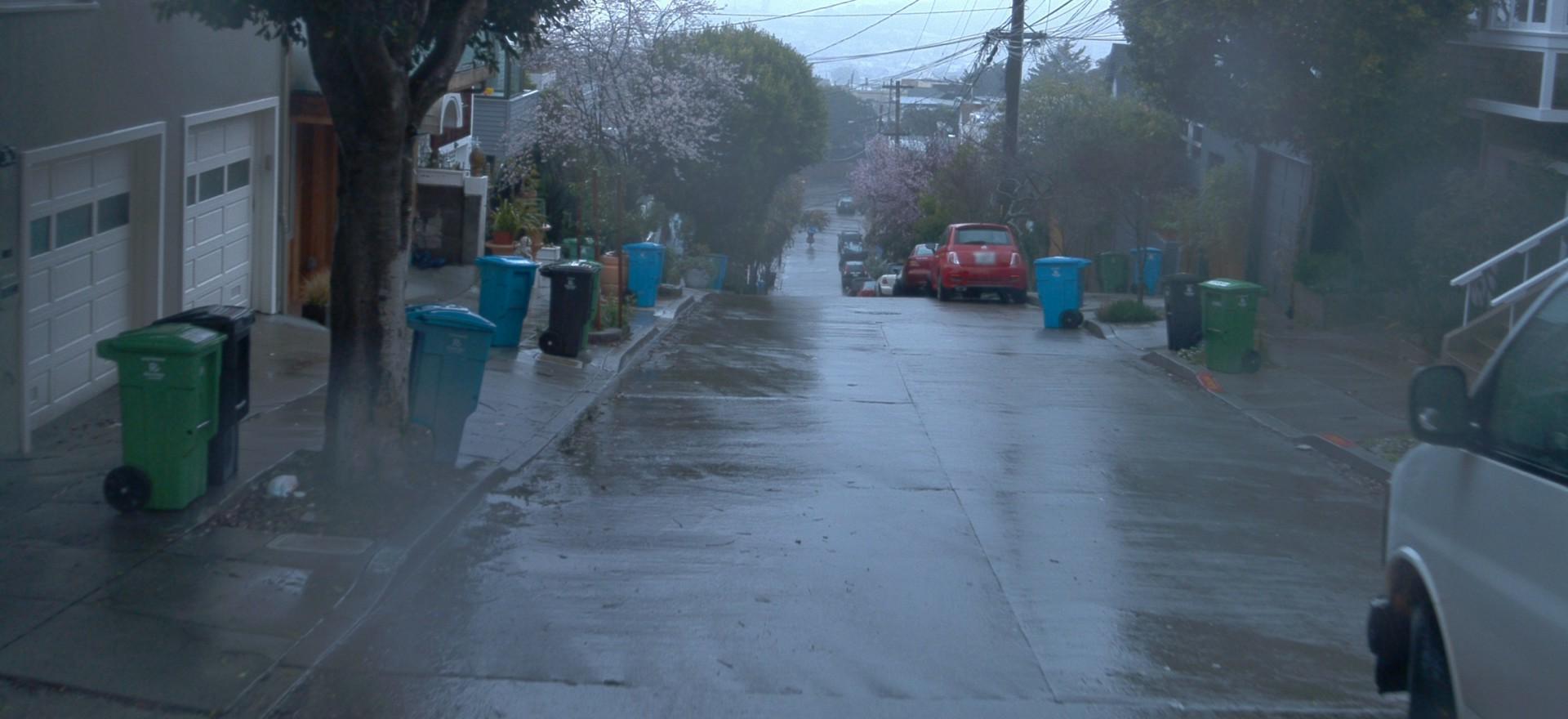}&
\includegraphics[width=0.24\linewidth]{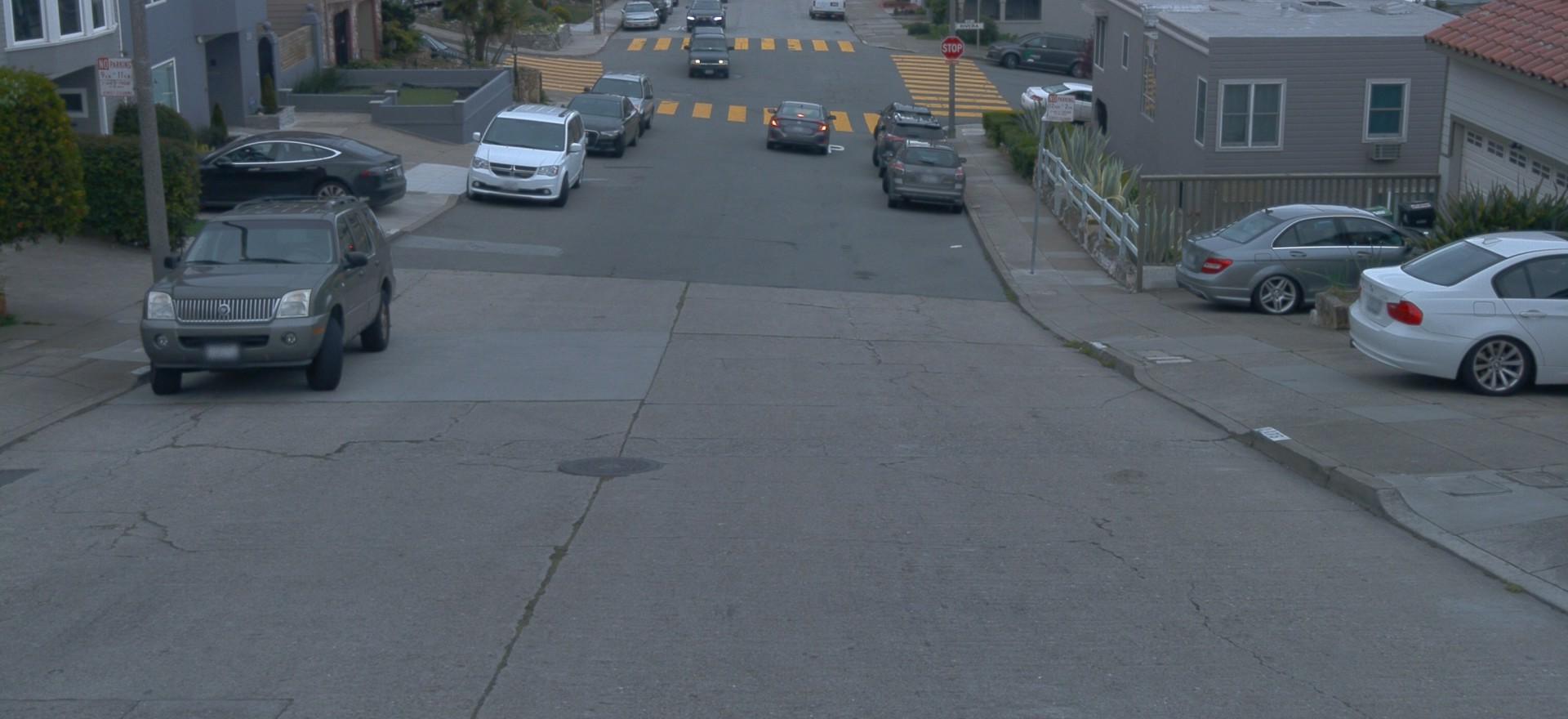}&
\includegraphics[width=0.24\linewidth]{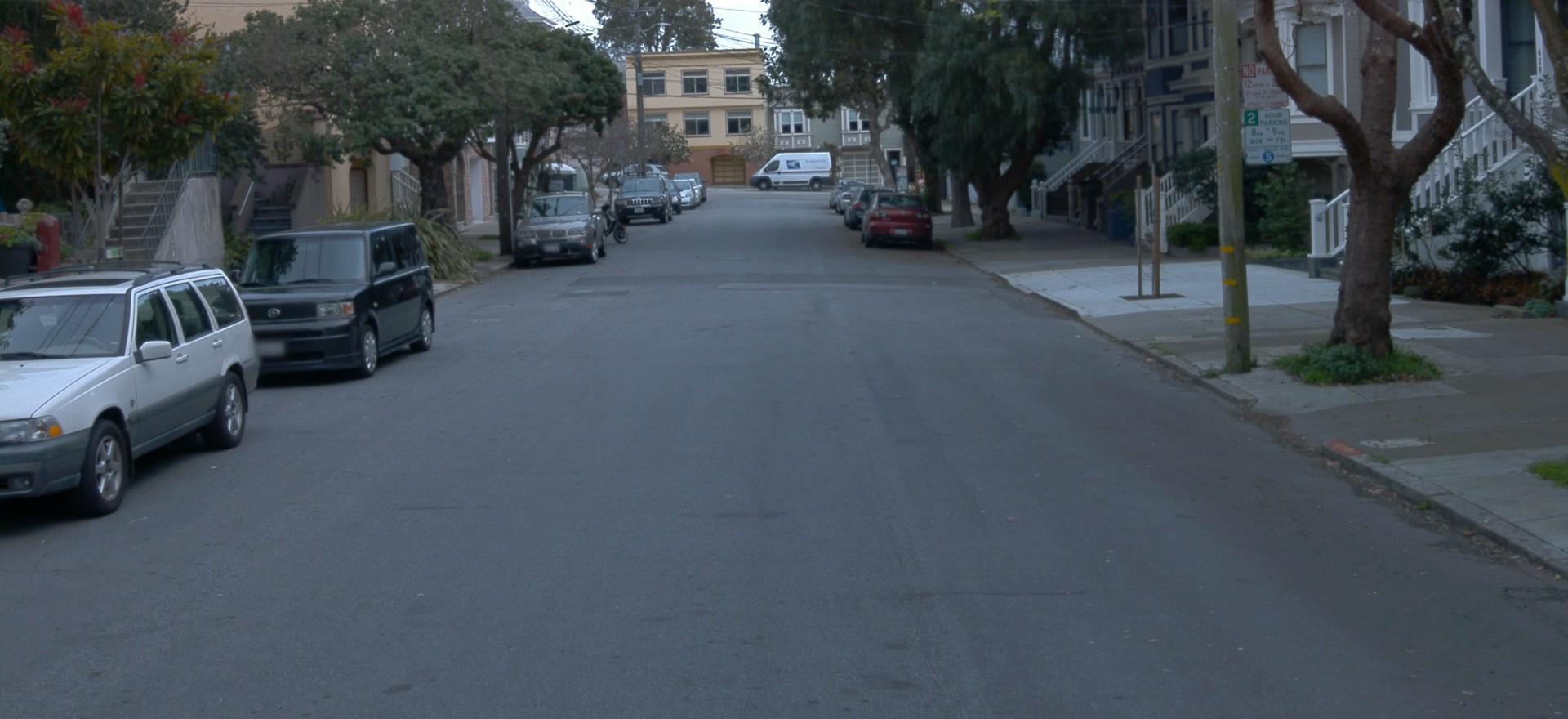}\\

\rotatebox{90}{\small  \hspace{7mm} GT}
\includegraphics[width=0.24\linewidth]{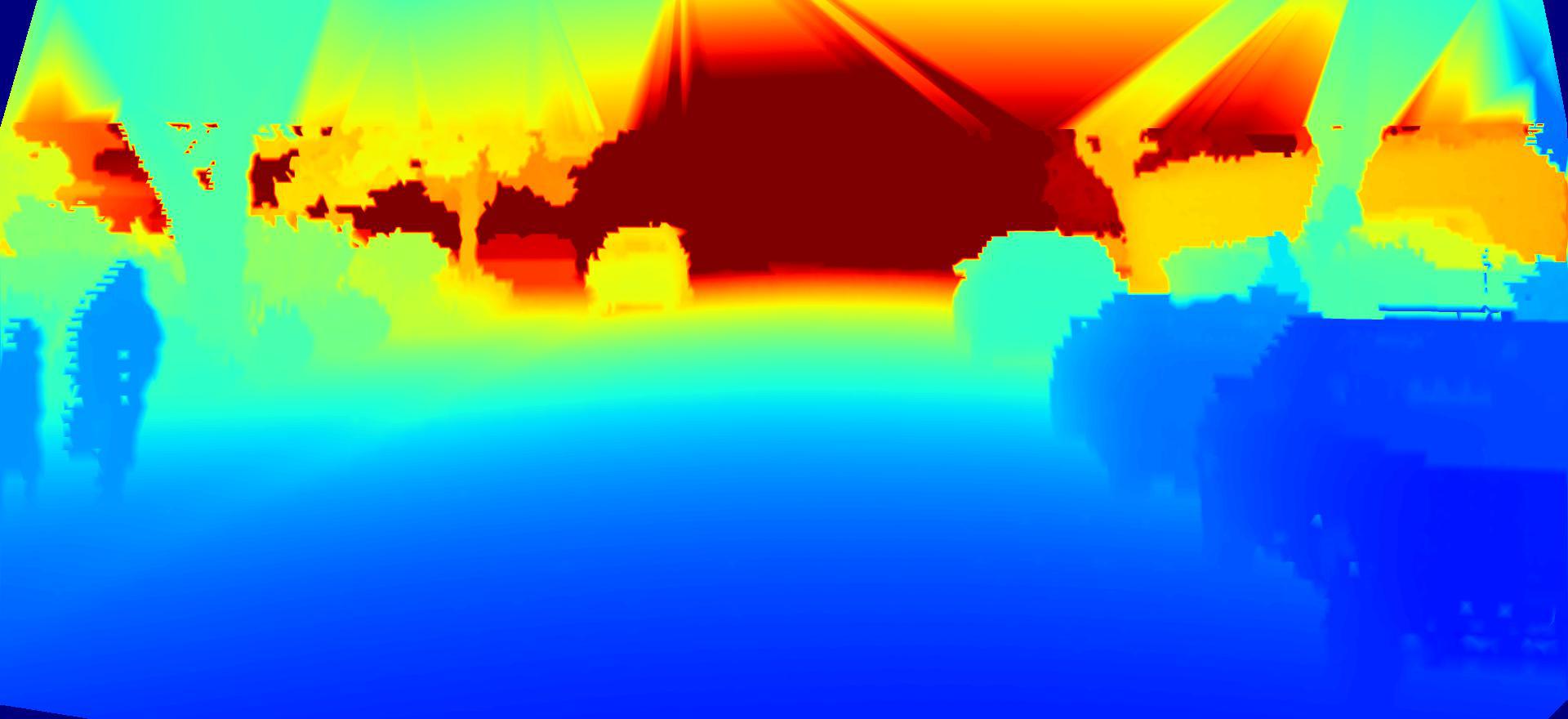}&
\includegraphics[width=0.24\linewidth]{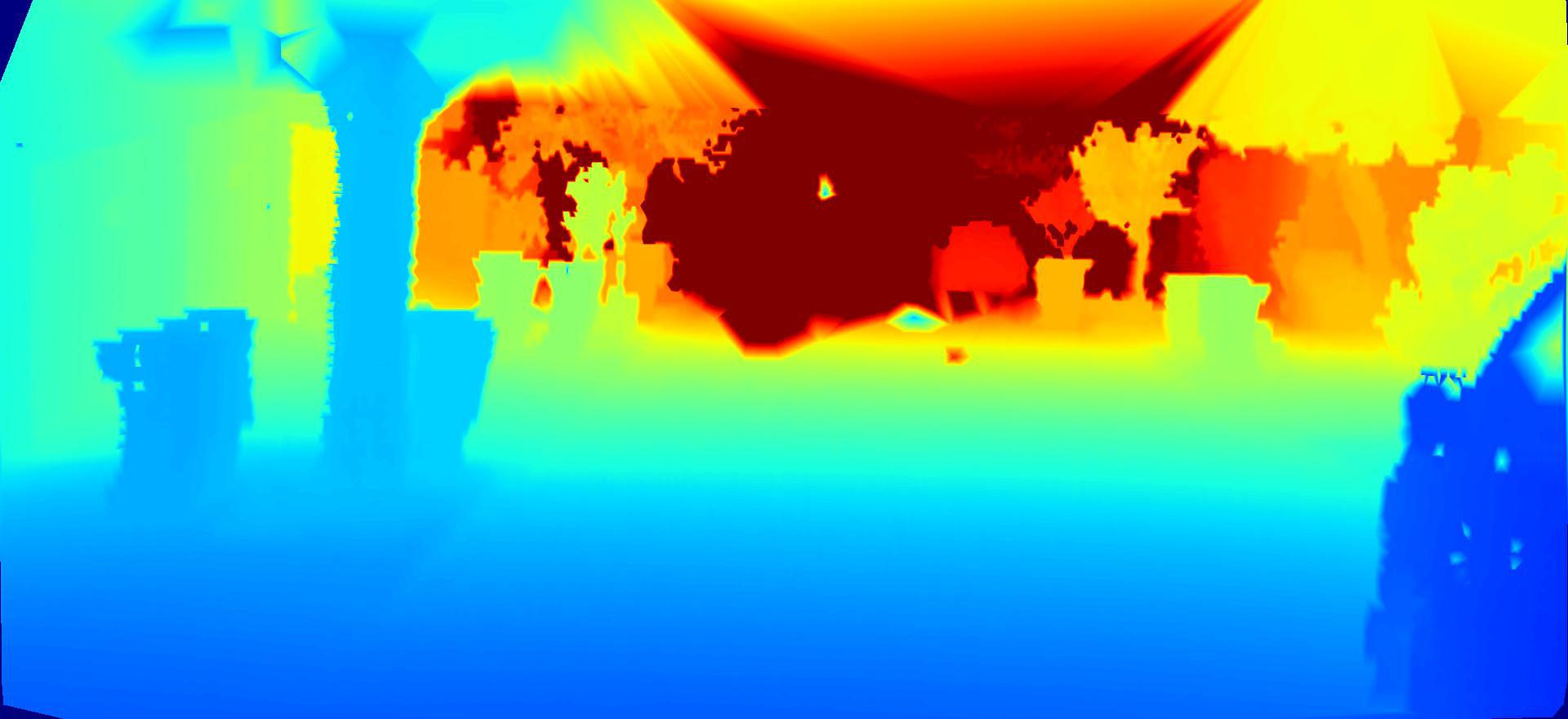}&
\includegraphics[width=0.24\linewidth]{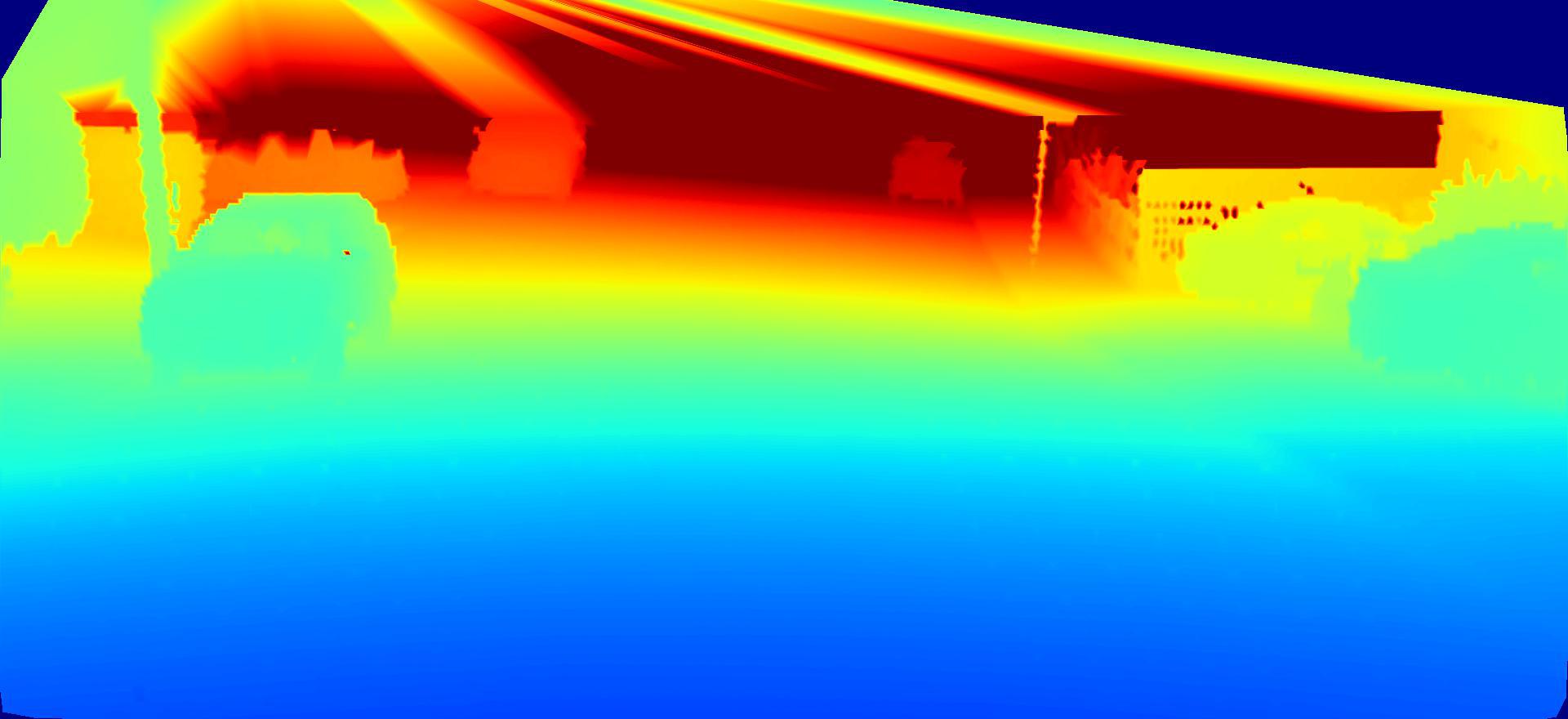}&
\includegraphics[width=0.24\linewidth]{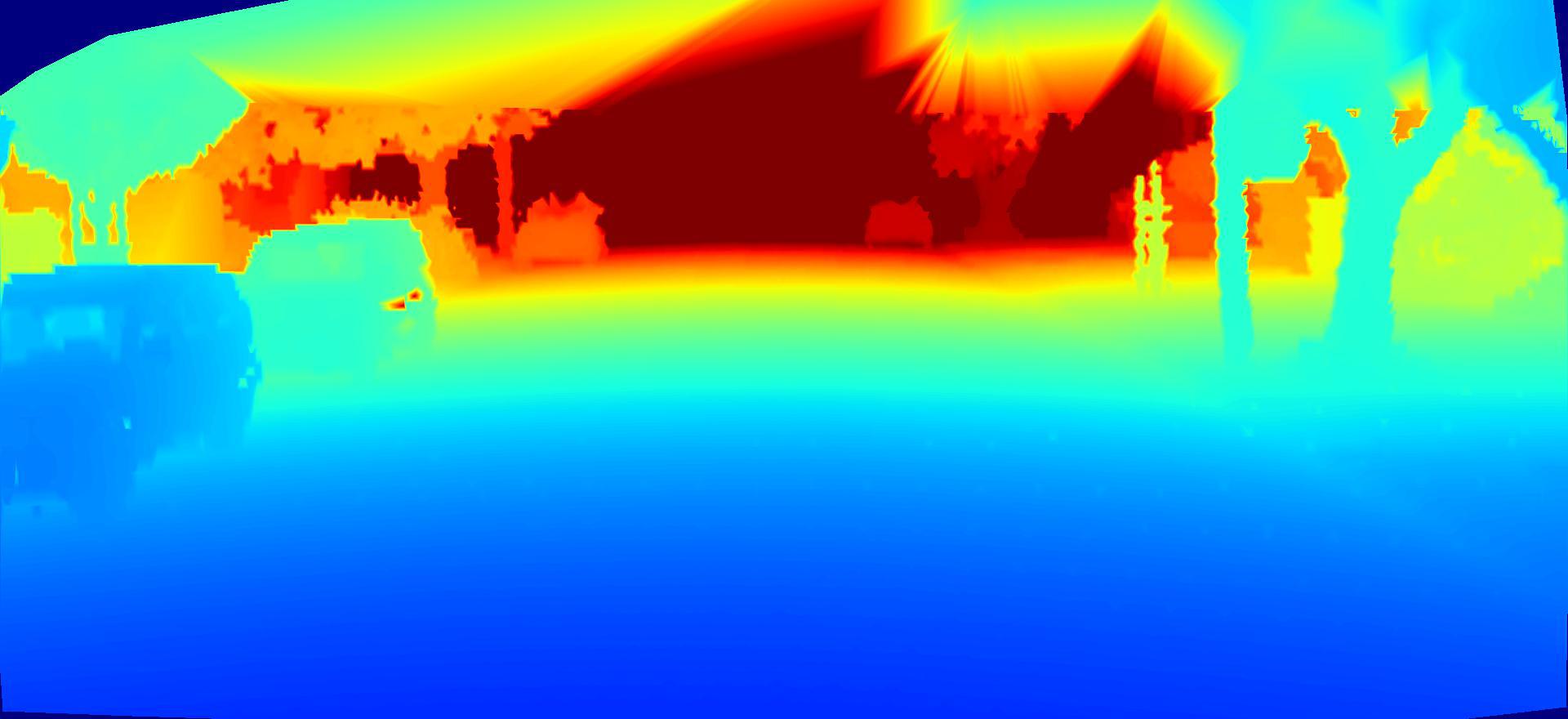}\\

\rotatebox{90}{\small  \hspace{5mm} DORN}
\includegraphics[width=0.24\linewidth]{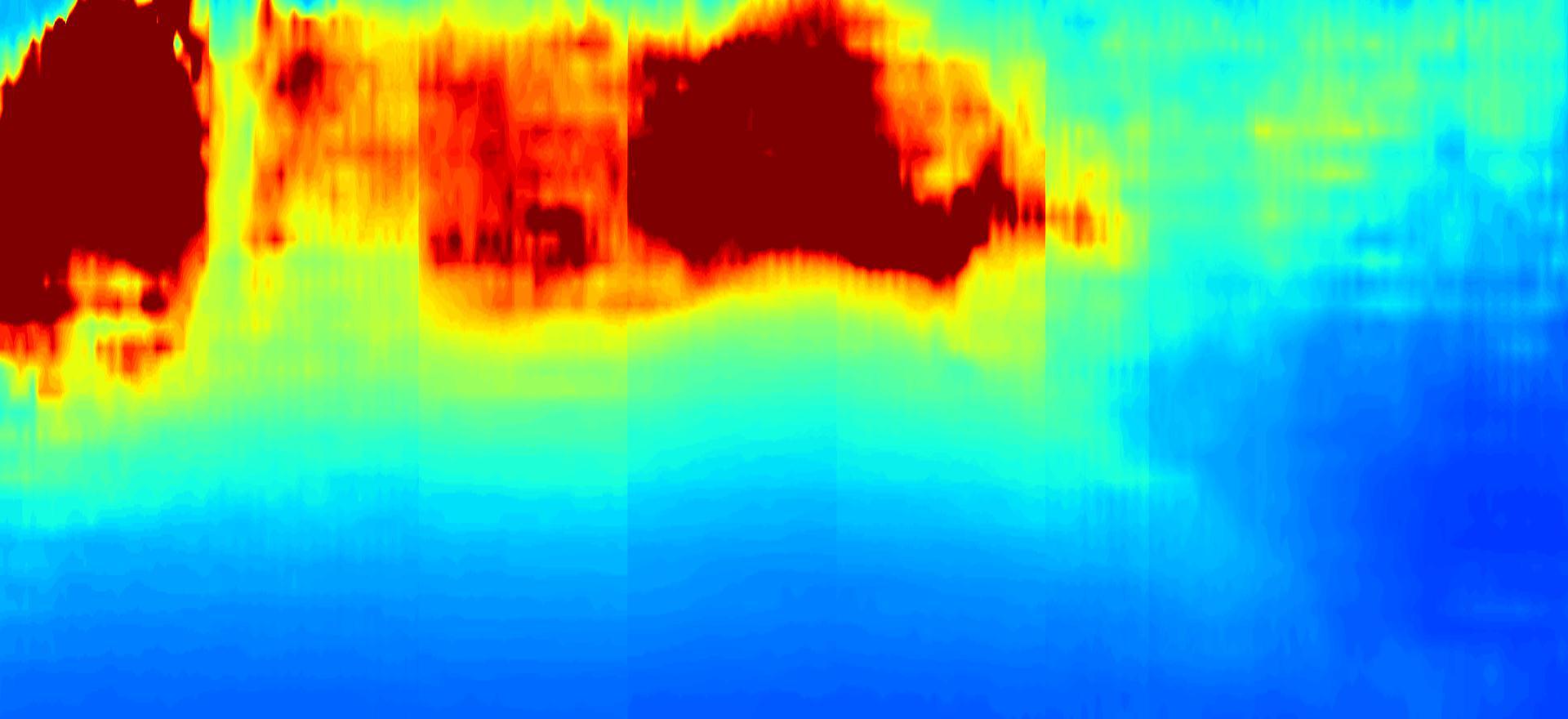}&
\includegraphics[width=0.24\linewidth]{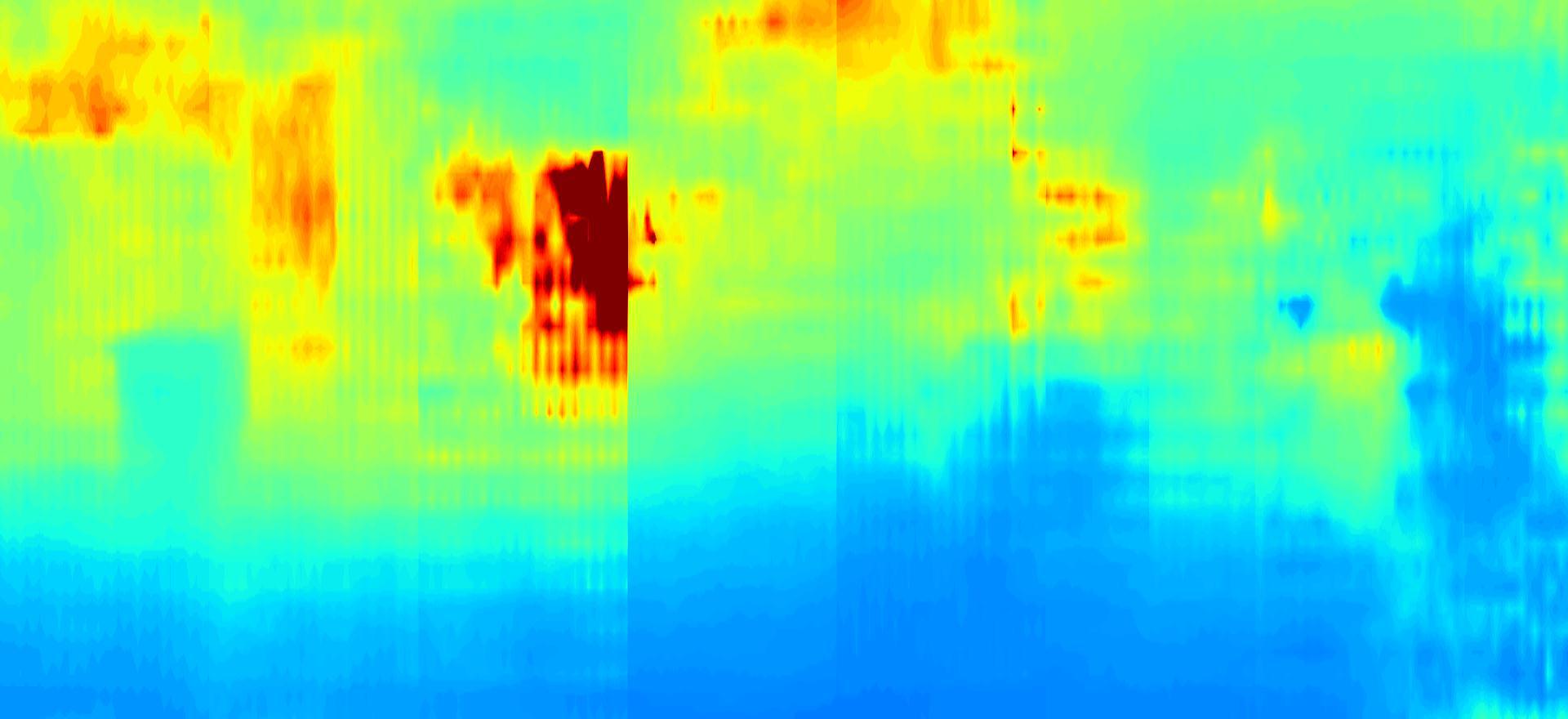}&
\includegraphics[width=0.24\linewidth]{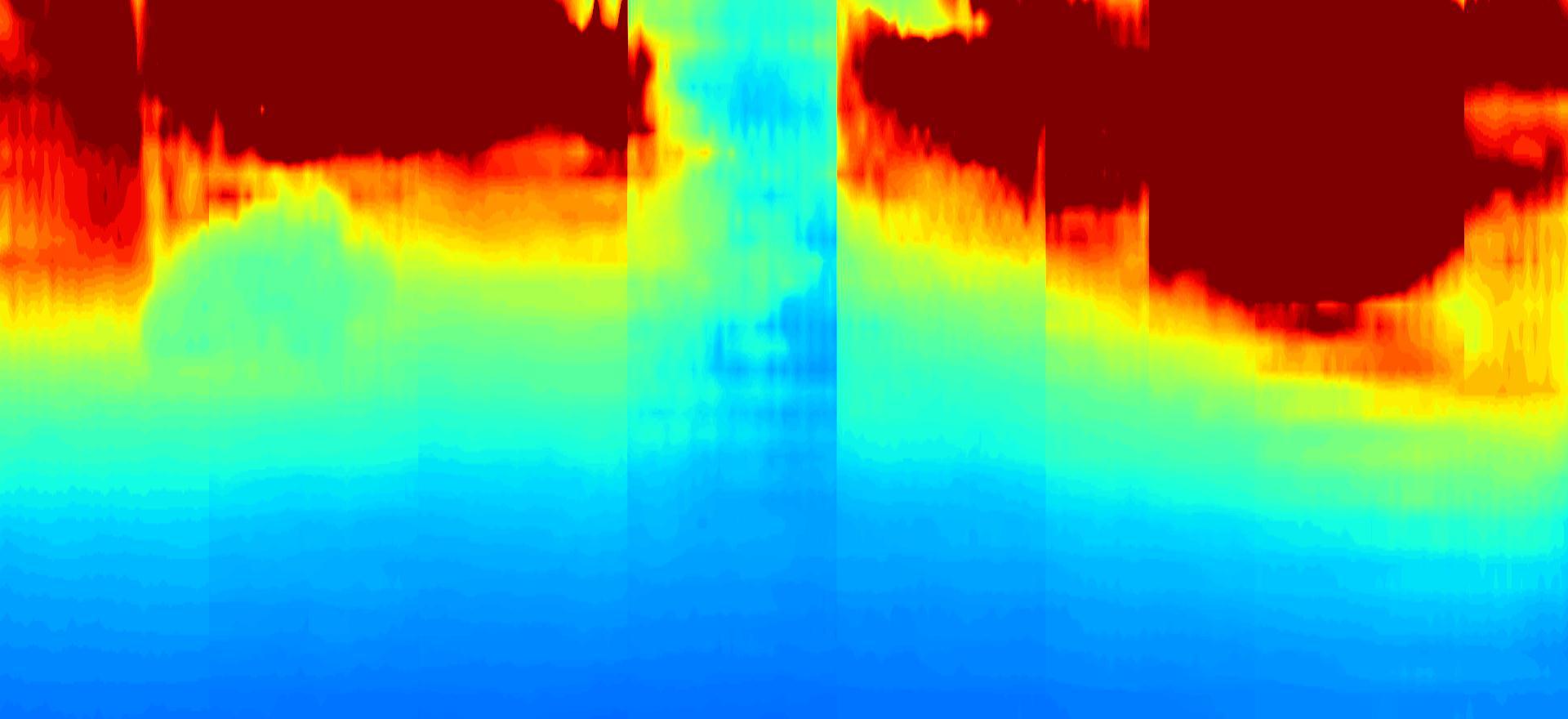}&
\includegraphics[width=0.24\linewidth]{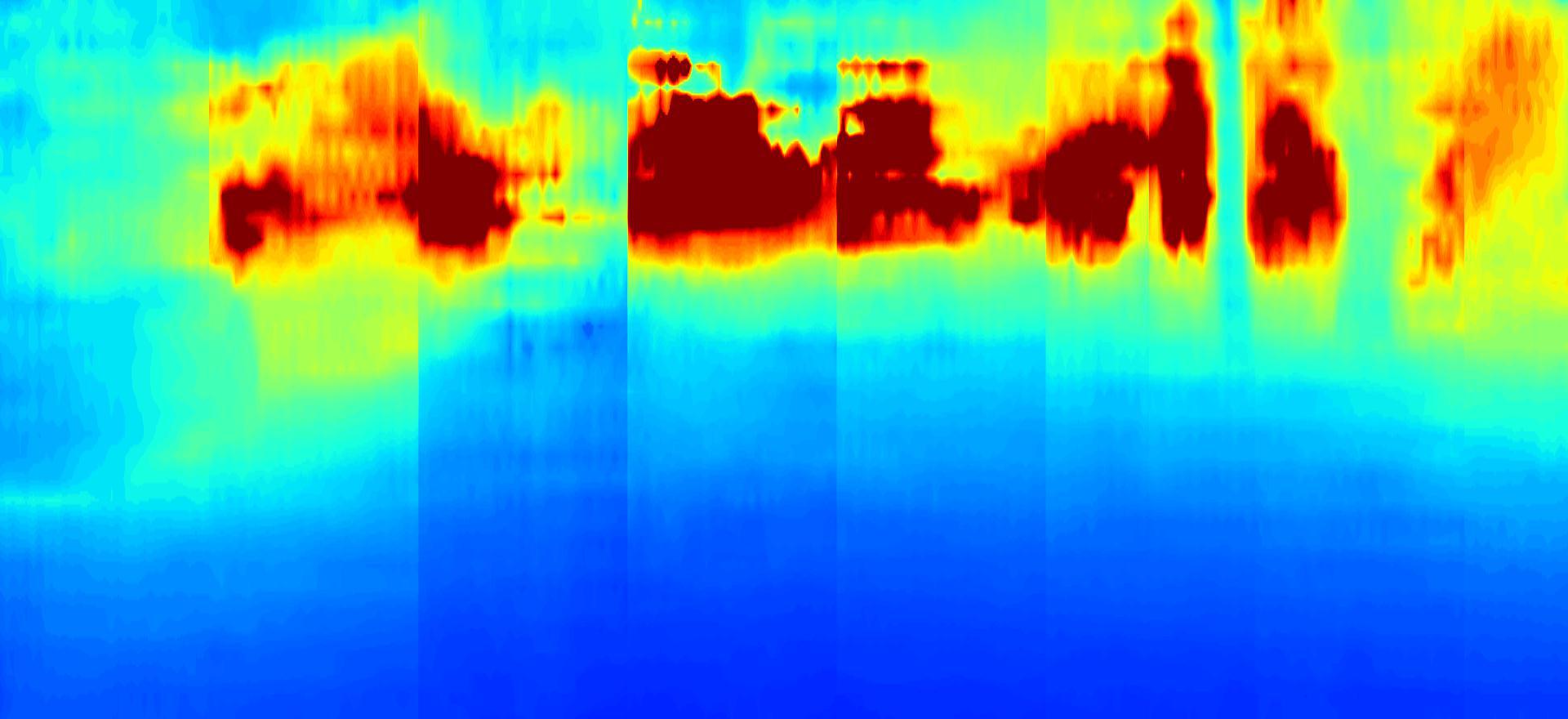}\\

\rotatebox{90}{\small  \hspace{0mm} NeuralRGBD}
\includegraphics[width=0.24\linewidth]{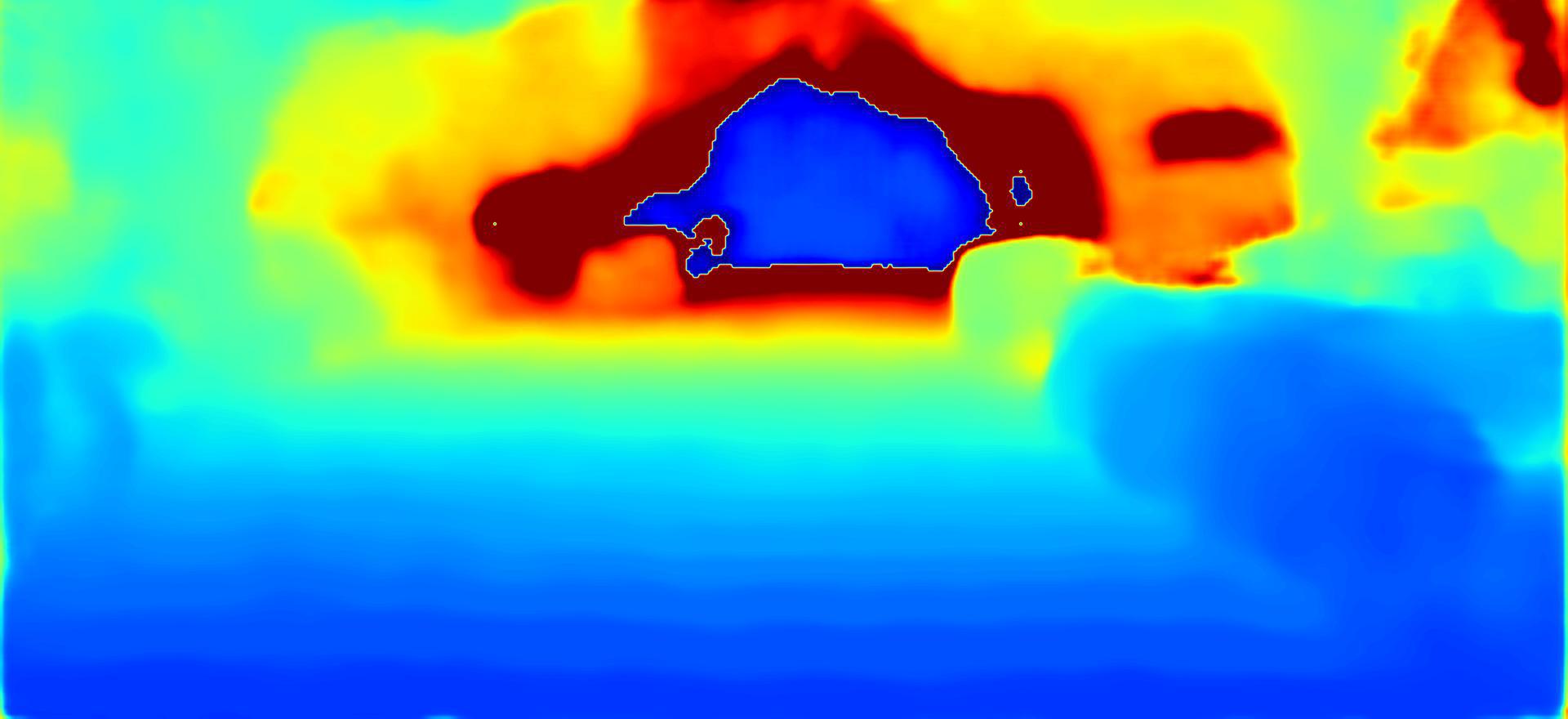}&
\includegraphics[width=0.24\linewidth]{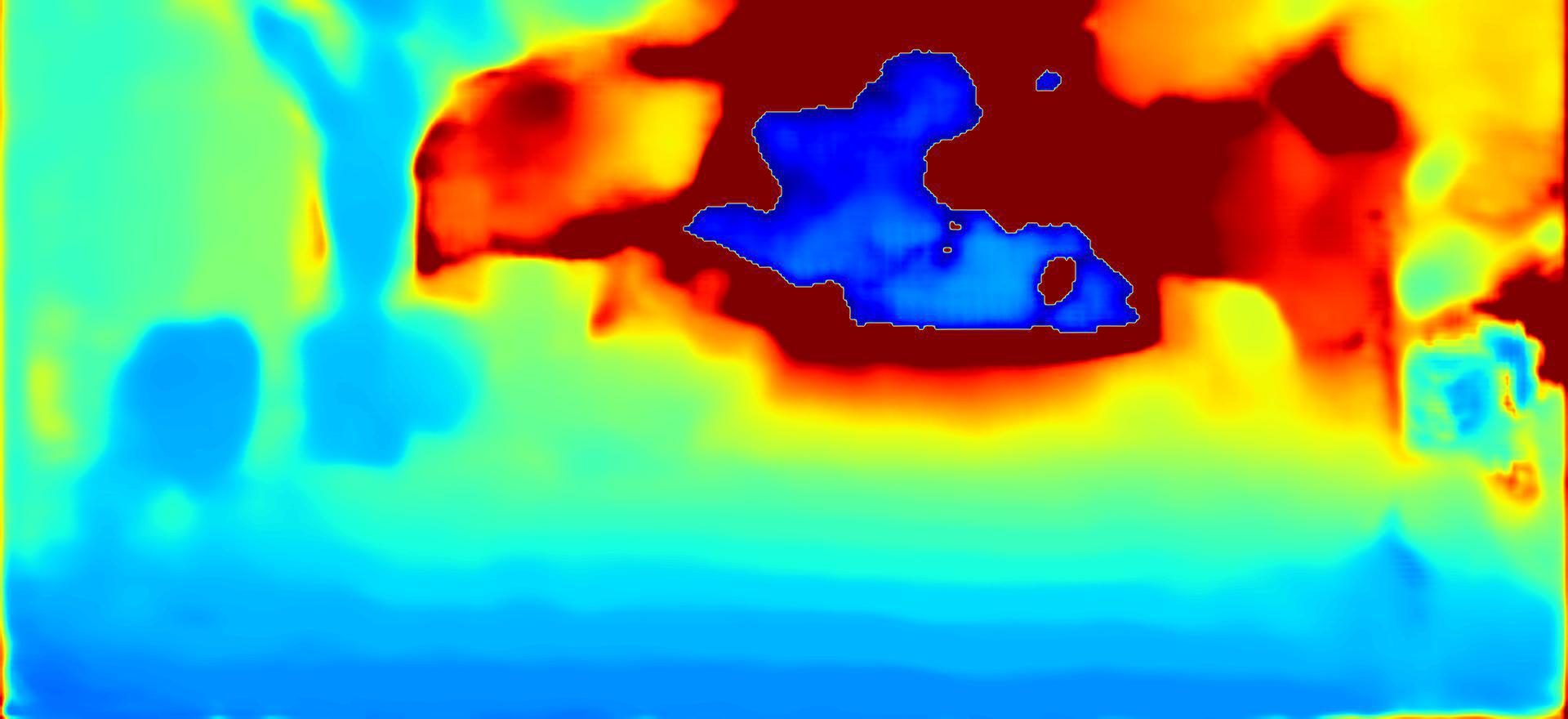}&
\includegraphics[width=0.24\linewidth]{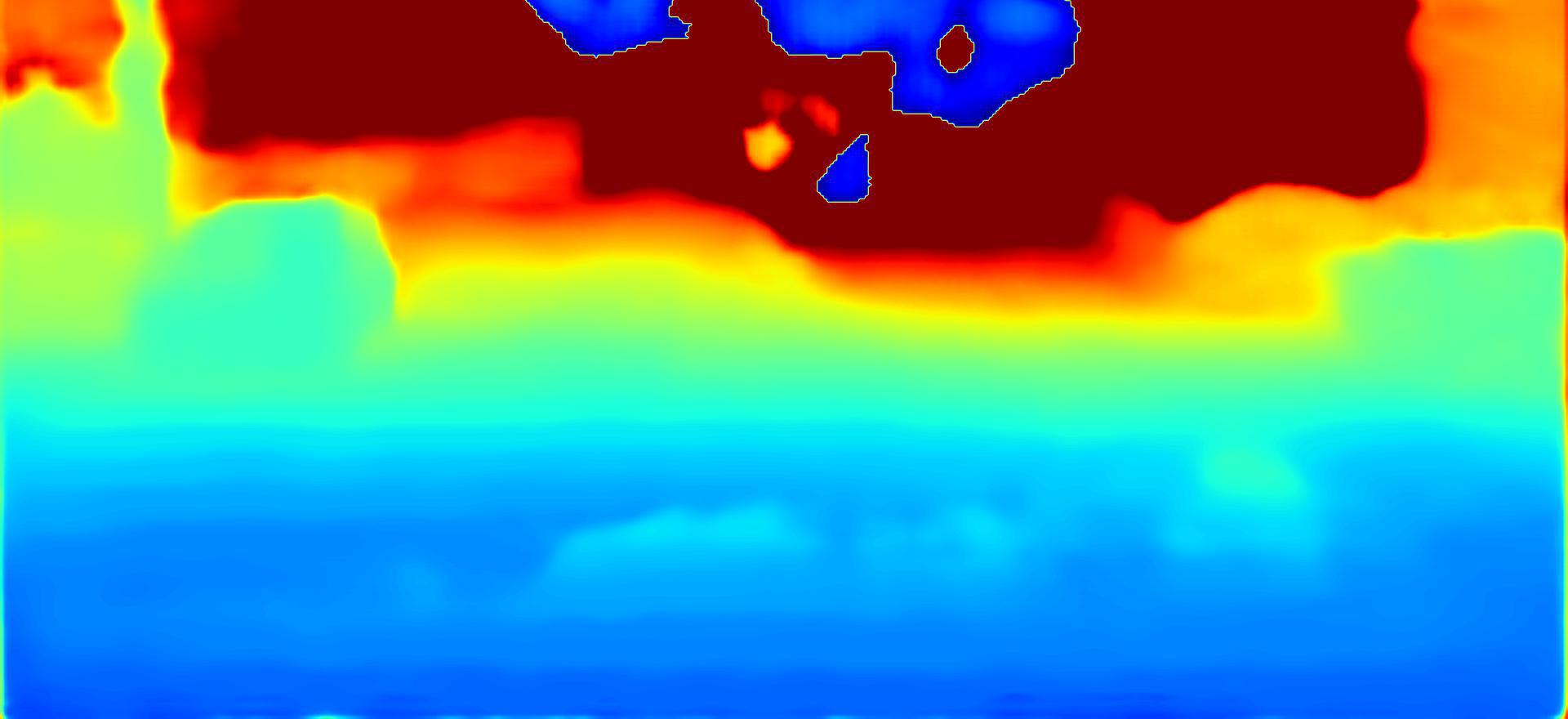}&
\includegraphics[width=0.24\linewidth]{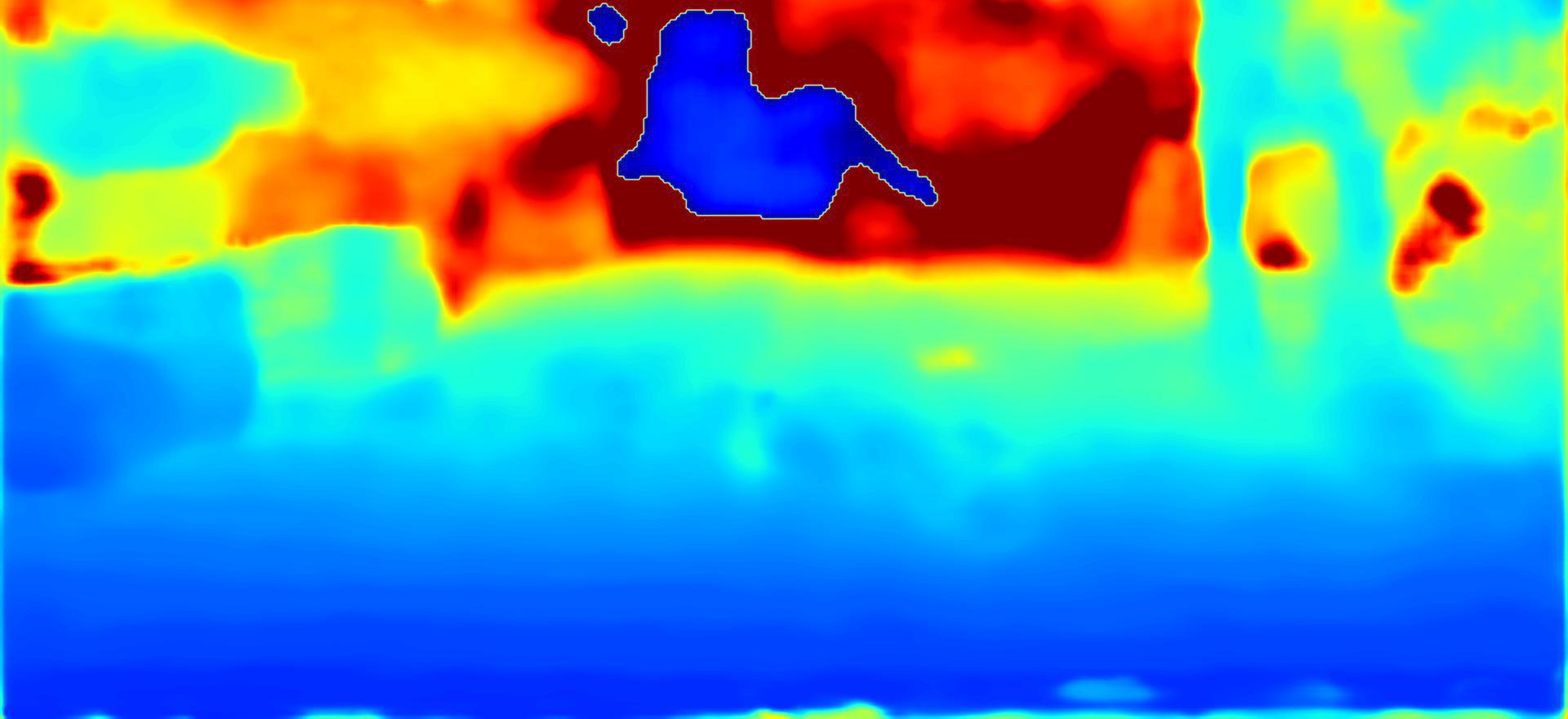}\\

\rotatebox{90}{\small  \hspace{6mm} Ours}
\includegraphics[width=0.24\linewidth]{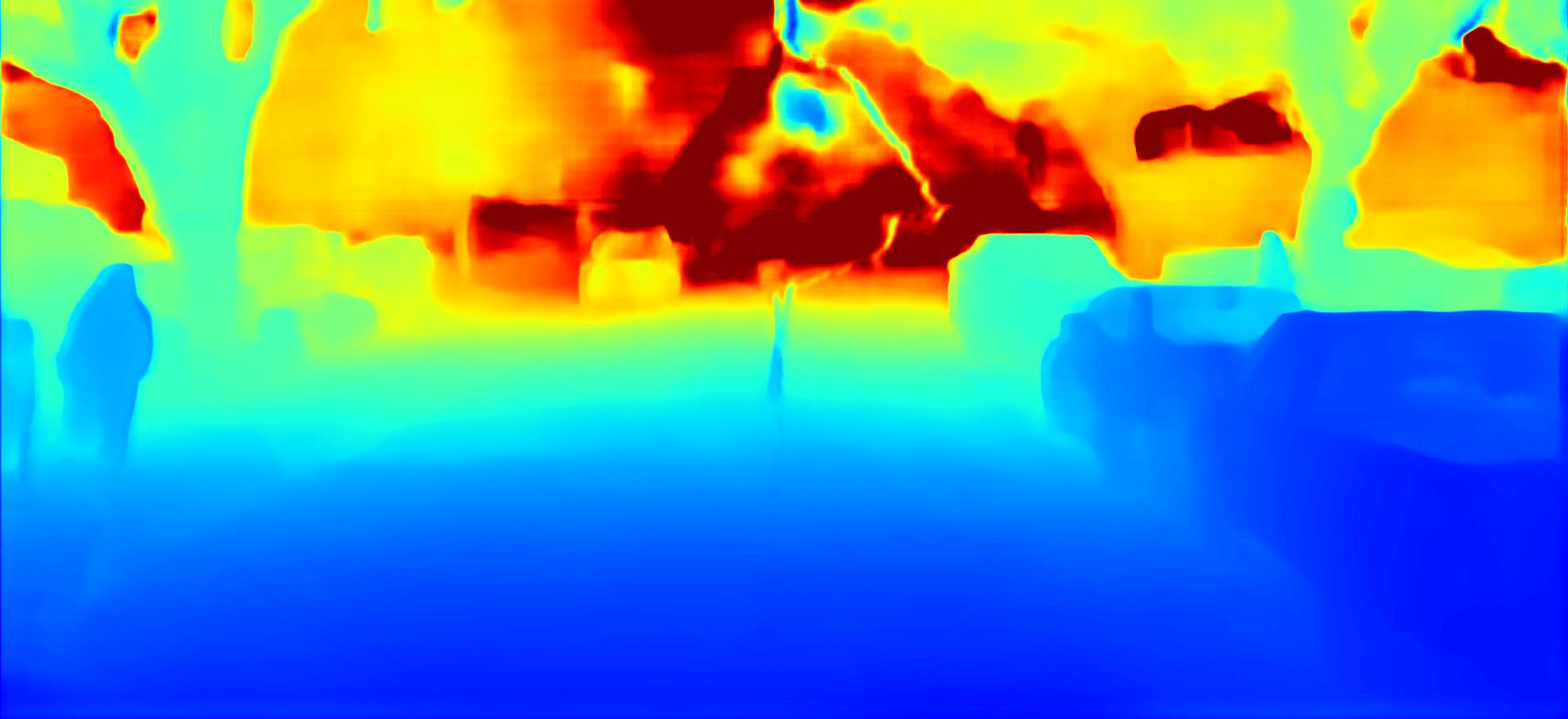}&
\includegraphics[width=0.24\linewidth]{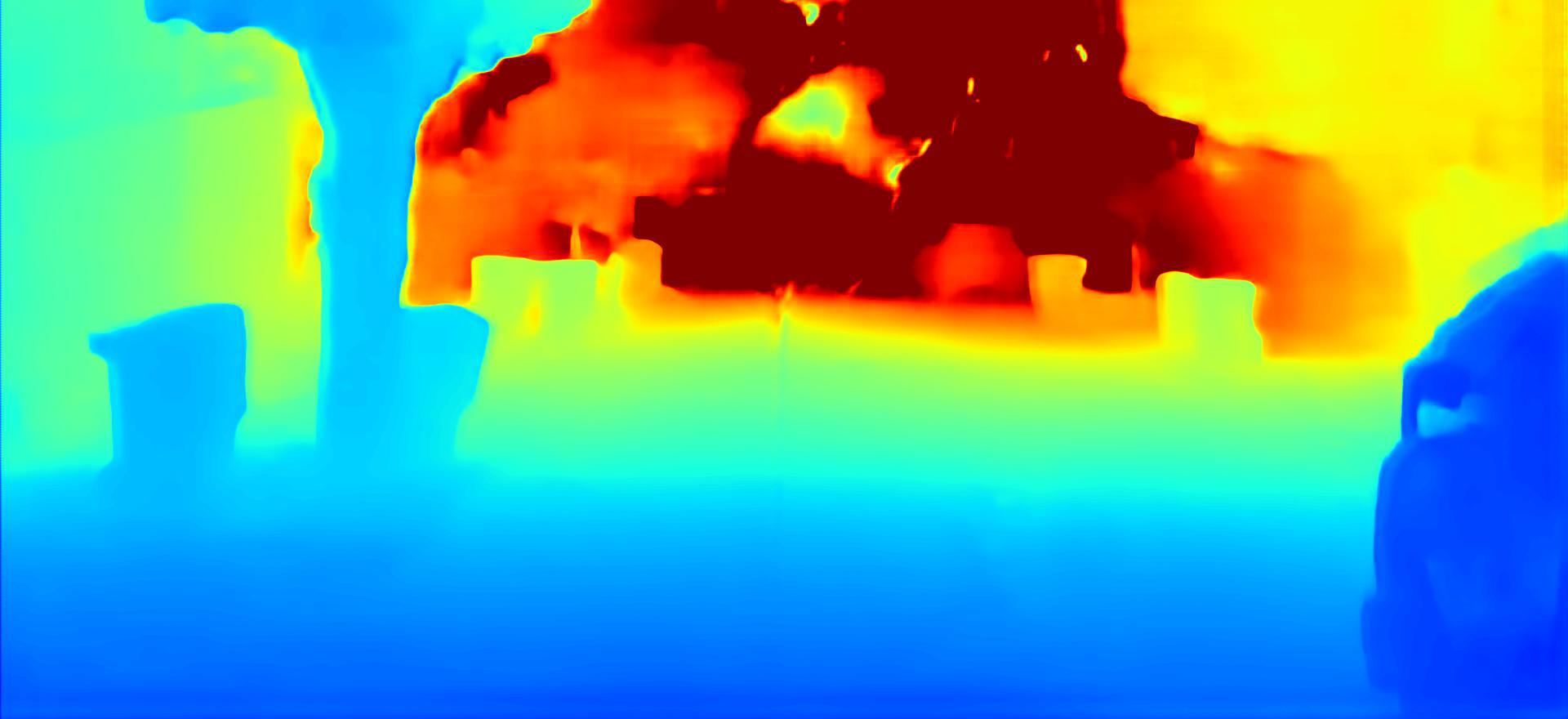}&
\includegraphics[width=0.24\linewidth]{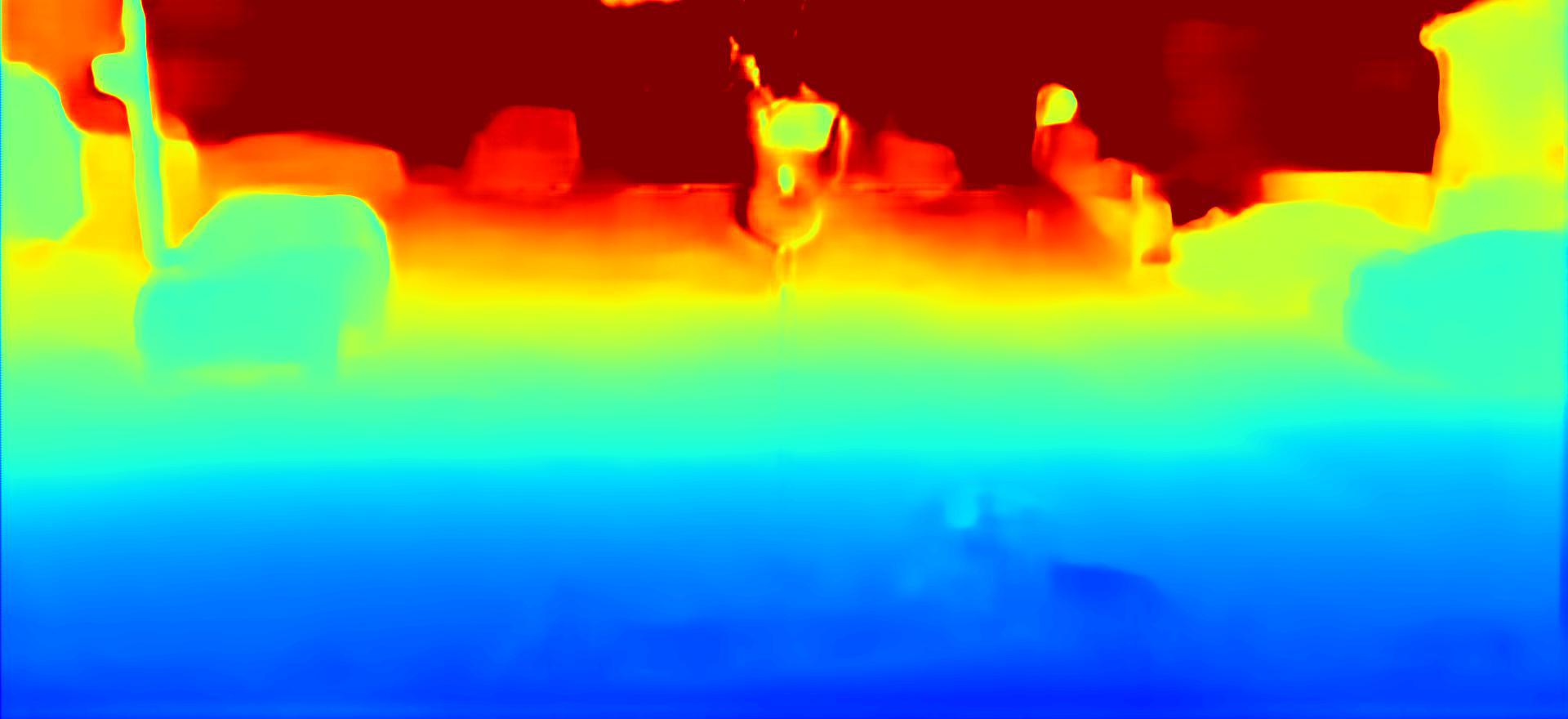}&
\includegraphics[width=0.24\linewidth]{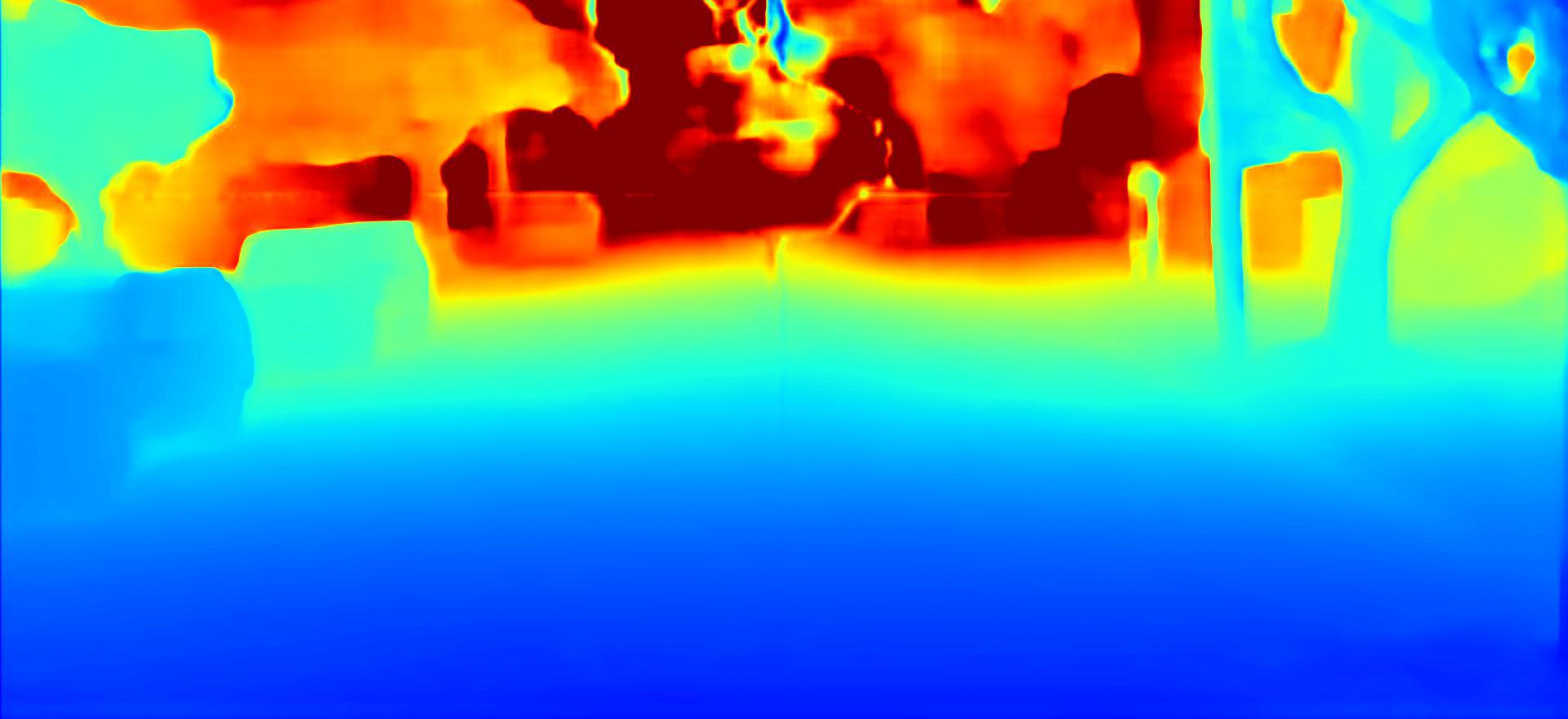}\\
\end{tabular}
\caption{ Cross-dataset comparisons between DORN~\cite{FuCVPR18-DORN}, NeuralRGBD~\cite{Liu_2019_CVPR}, and ours on the Waymo dataset. All the depth maps are inferenced by models trained on KITTI.}
\label{fig:Waymo results}
\end{figure*}

\subsection{Results}
We conduct extensive experiments to evaluate the performance of our method and state-of-the-art methods. Our method is able to produce more accurate depth maps and outperforms the contemporaneous methods on most evaluation metrics. In addition, our method is more robust and shows great generalization ability.

\paragraph{Quantitative Evaluation}
On the KITTI dataset, we train our model in the Eigen split, and the Uhrig split separately. Table \uppercase\expandafter{\romannumeral1} summarizes the quantitative evaluation results of our method and other state-of-the-art baselines in both splits. For a fair comparison, we use exactly the same evaluation code provided by Zhou et al.~\cite{Zhou2017} to evaluate all the methods except Eigen et al.~\cite{EgienPF14}. We directly use the results reported on Eigen et al.~\cite{EgienPF14} because the provided source code only produces low-resolution $28\times 144$ or $27\times 142$ depth maps, but we evaluate on full-resolution depth maps. The results are much worse if we upsample their output low-resolution depth maps.

Regarding the metrics, we include widely used ones from previous work~\cite{YinShi2018, FuCVPR18-DORN}, and metrics used by the KITTI single image depth estimation benchmark. They are abs rel: absolute relative error; sq rel: square relative error; rms: root mean square; log rms: log root mean square; irmse: inverse root mean square error; SIlog: scale-invariant logarithmic error; $\delta_i$: the percentage of pixels with relative depth error $\delta < 1.25^i$. The $\downarrow$ indicates the lower the better, the $\uparrow$ does the opposite. 

As shown in the Table \uppercase\expandafter{\romannumeral1}, our method outperforms state-of-the-art methods in both splits. In the Eigen split, our method has outperformed several state-of-the-art depth estimation methods by a large margin. In the Uhrig split, our model additionally takes the depth maps generated by NeuralRGBD~\cite{Liu_2019_CVPR} for depth fusion, and has about 20-30\% improvement in most metrics. 

Table \uppercase\expandafter{\romannumeral2} compares our method with two representative approaches on the ScanNet dataset. As shown in Table \uppercase\expandafter{\romannumeral2}, our method performs better on the first nine metrics and achieve comparable performance with DORN~\cite{FuCVPR18-DORN} on metric $\delta_3$. The depth proposals we used are the same as the model in the KITTI Uhrig split. Besides depth proposals generated by the flow-to-depth layer, the result of NeuralRGBD~\cite{Liu_2019_CVPR} serves as a depth proposal on this model, which speeds up the training process and improves performance.

\begin{table}
\renewcommand\arraystretch{1.6}
\begin{center}
\caption{Quantitative evaluation of ablation study}
\begin{tabular}{@{}l@{\hspace{3mm}}l@{\hspace{3mm}}c@{\hspace{3mm}}c@{\hspace{3mm}}c@{\hspace{3mm}}c@{\hspace{3mm}}c@{\hspace{3mm}}c@{}}
\hline
Method & abs rel $\downarrow$ & sq rel $\downarrow$ & rms $\downarrow$ & SIlog $\downarrow$ & $\delta_1 \uparrow$ \\
\hline
RGB frames only&0.120&0.817&4.690&0.189&0.858\\
Ours (w/o refinement)&0.085&0.522&3.767&0.148&0.906\\
Ours (w/ refinement)&\textbf{0.081}&\textbf{0.488}&\textbf{3.651}&\textbf{0.144}&\textbf{0.912}\\
\hline
\end{tabular}
\end{center}
\label{table:ablation study}
\end{table}

\paragraph{Qualitative evaluation}

Fig.~\ref{fig:KITTI} illustrates some qualitative results. Let us look at the green box in Scene 1 and Scene 2:  NeuralRGBD~\cite{Liu_2019_CVPR} misses the top of a van behind two cars in Scene 1, and only estimates the bottom part of a truck in Scene 2, which means image priors are not used properly in these areas. Thus, this phenomenon indicates that our method can take advantage of image priors when the geometrical constraints are not reliable.

Then look at the red box in Scene 2 and Scene 3. DORN~\cite{FuCVPR18-DORN} produces a blurry depth map that can not differentiates object boundaries, but NeuralRGBD~\cite{Liu_2019_CVPR} and our method produce reasonably sharper results. Note that a common characteristic of NeuralRGBD~\cite{Liu_2019_CVPR} and ours is that we both use geometrical information. 

Fig.~\ref{fig:scanNet results} shows the comparisons on the ScanNet dataset, where the first row shows depth maps and the second row shows error maps. As show in the error maps, we produce depths with lower error compared to NeuralRGBD~\cite{Liu_2019_CVPR} and DORN~\cite{FuCVPR18-DORN}. Our output depth map is less noisy and more complete.

\begin{figure}
\centering
\begin{tabular}{@{}c@{\hspace{1mm}}c@{\hspace{1mm}}c@{}}
&Scene 1&Scene 2\\
\rotatebox{90}{\small \hspace{3.2mm} $I_t$}&
\includegraphics[width=0.45\linewidth]{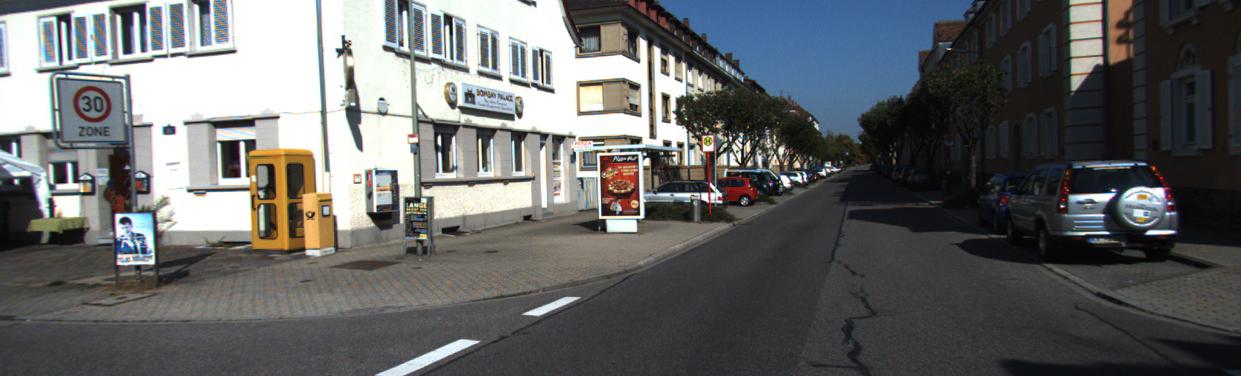}&
\includegraphics[width=0.45\linewidth]{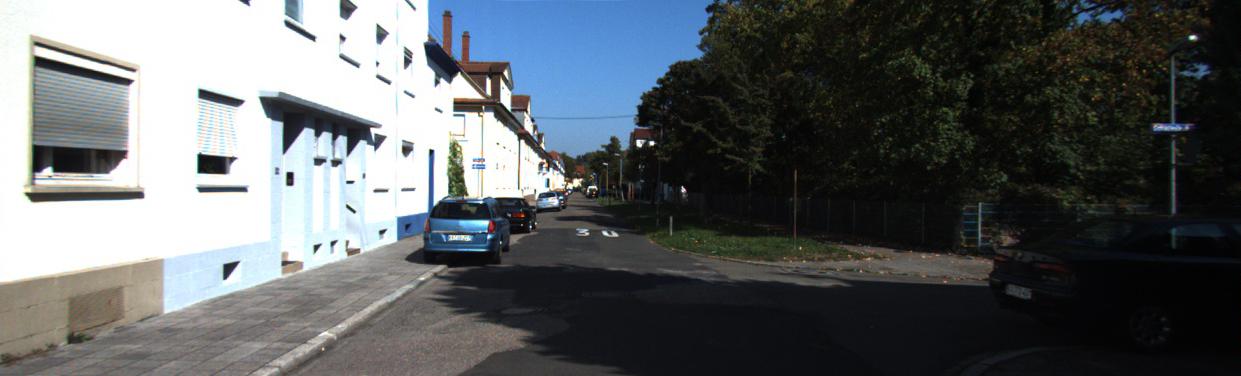}\\

\rotatebox{90}{\small \hspace{0.2mm} Without}&
\includegraphics[width=0.45\linewidth]{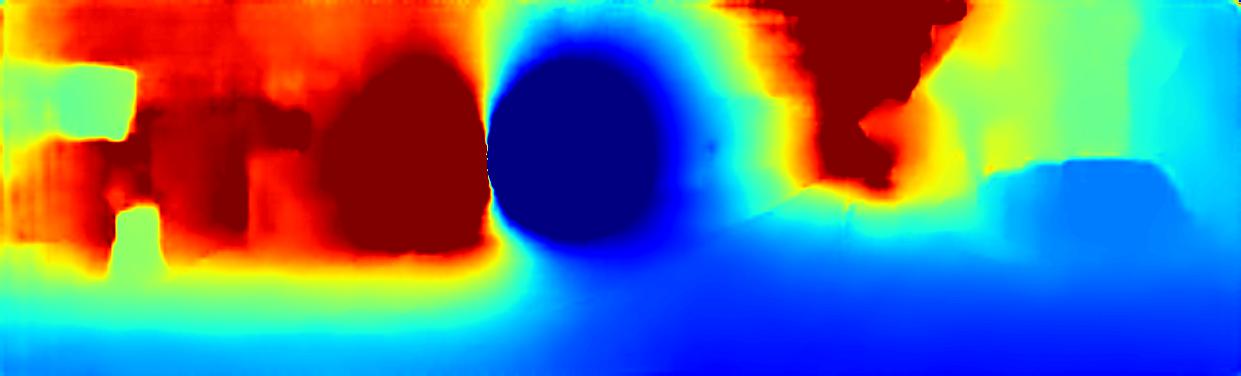}&
\includegraphics[width=0.45\linewidth]{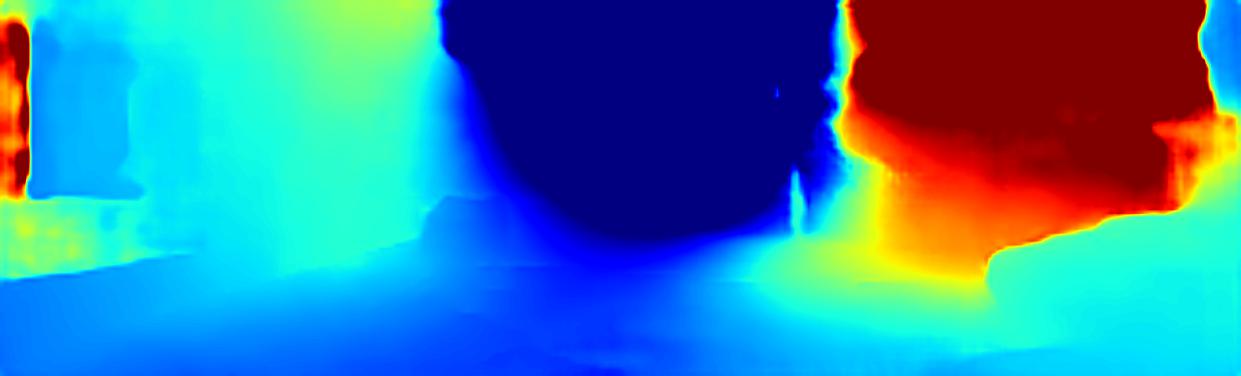}\\

\rotatebox{90}{\small \hspace{1.7mm} With}&
\includegraphics[width=0.45\linewidth]{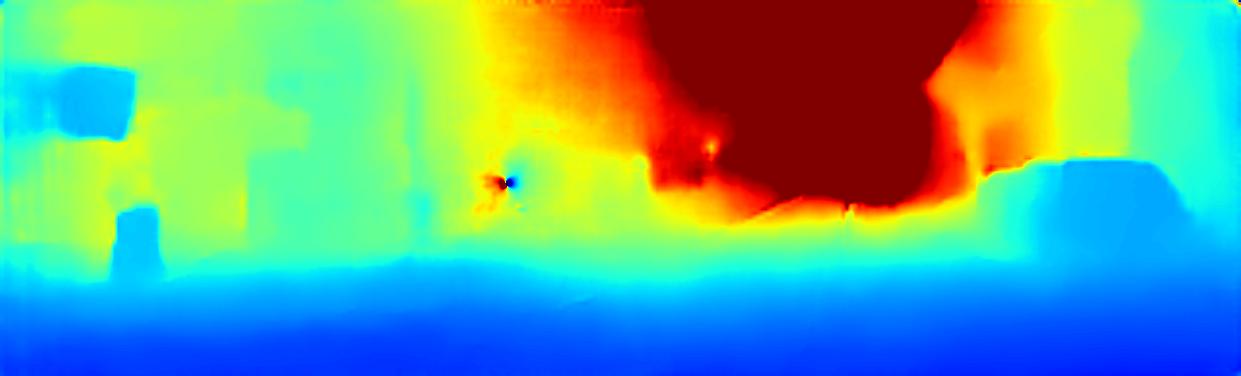}&
\includegraphics[width=0.45\linewidth]{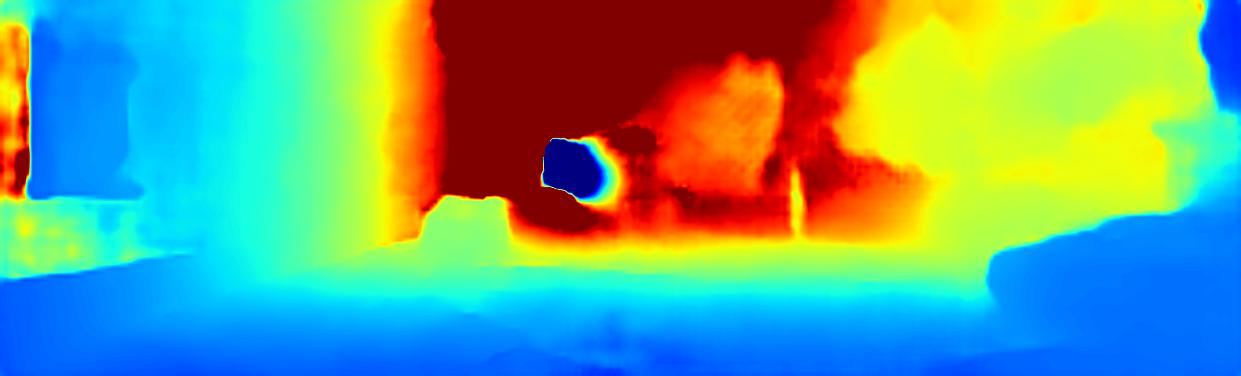}\\

\end{tabular}
\vspace{2mm}
\caption{Visualization of depth proposals with (the third row) and without (the second row) camera pose refinement.}
\label{fig:Pose refinement}
\end{figure}

\paragraph{Ablation study}
The accuracy of relative camera poses can significantly affect the video-based depth estimation performance. Fig.~\ref{fig:Pose refinement} shows depth proposals generated with and without pose refinement in two extreme examples. In the second row, without pose refinement, the initial camera pose produces poor depth proposals that have a vast region of negative depths. After pose refinement, in the third row, we can get depth proposals with higher confidence. We show a quantitative comparison between models with and without pose refinement in Table \uppercase\expandafter{\romannumeral 4}, and camera pose refinement can give us an improvement about 3 to 6 percents on these metrics.

We also have an ablation experiment by training the depth fusion network to estimate depth directly from the target frame and source frames. The results of this experiment are shown in the first row of Table \uppercase\expandafter{\romannumeral 4}. Our complete model performs much better than the ablated model. This comparison validates the strength of the flow-to-depth layer in our model.

\paragraph{Cross dataset evaluation}
Table \uppercase\expandafter{\romannumeral 3} reports the quantitative results of cross dataset evaluation on the Waymo dataset. Our model (trained on KITTI and test on Waymo) suffers less performance degeneration than NeuralRGBD~\cite{Liu_2019_CVPR}, DORN~\cite{FuCVPR18-DORN}, and SfMLeaner~\cite{Zhou2017} in cross dataset evaluation. Fig.~\ref{fig:Waymo results} shows the cross dataset visual results of our model and three baselines. These visual results suggest that our depth proposals can often preserve object boundaries in the estimated depth maps, even on the cross dataset results. 


\section{Conclusion}

We have presented a video depth estimation method that builds upon a novel flow-to-depth layer. This layer can help refine camera poses and generate depth proposals. Beyond the depth proposals computed from the flow-to-depth layer, depth maps estimated by other methods can also serve as depth proposals in our model. In the end, a depth fusion network fuses all depth proposals to generate a final depth map. The experiments show that our presented model outperforms all other state-of-the-art depth estimation methods on the KITTI dataset, ScanNet dataset, and shows excellent generalization ability on the Waymo dataset. We hope our model can be a practical tool for other researchers and inspire more future work on monocular video depth estimation.

\normalsize	
\vspace{2mm}
\bibliographystyle{IEEEtran}
\bibliography{egbib}

\end{document}